\pdfoutput=1
\documentclass[11pt]{article}

\usepackage[]{acl}

\usepackage{times}
\usepackage{latexsym}

\usepackage[T1]{fontenc}
\usepackage[utf8]{inputenc}

\usepackage{microtype}

\usepackage{latexsym}
\usepackage{url}
\usepackage{dsfont}
\usepackage{acronym}
\usepackage[noend]{algorithmic}
\usepackage{algorithm}
\usepackage{bbm}
\usepackage{beramono}
\usepackage{bm}
\usepackage{booktabs}
\usepackage[skip=2pt,belowskip=2pt,aboveskip=2pt]{caption}
\usepackage[T1]{fontenc}
\usepackage{graphicx}
\usepackage{hyperref}
\usepackage{listings}
\usepackage{multirow}
\usepackage{natbib}
\usepackage{subcaption}
\usepackage{tabularx}
\usepackage{enumerate}
\usepackage[inline]{enumitem}
\usepackage{numprint}
\usepackage[normalem]{ulem}
\usepackage{xspace}
\usepackage{graphicx}
\usepackage{xcolor}
\usepackage{colortbl}
\usepackage{amssymb}
\usepackage{amsmath}
\usepackage{cleveref}
\usepackage{multicol}
\usepackage{xtab}

\usepackage{latexsym}
\usepackage{float}

\usepackage{array}
\usepackage{makecell}
\usepackage{graphicx}
\usepackage{enumitem}

\newcommand{\OurDataset}{\textsc{GraphelSums}}

\title{Summarization with Graphical Elements}

\author{Maartje ter Hoeve \\
  University of Amsterdam \\
  \href{mailto:m.a.terhoeve@uva.nl}{m.a.terhoeve@uva.nl} \\\And
  Julia Kiseleva \\
  Microsoft Research \\
  \href{mailto:julia.kiseleva@microsoft.com}{julia.kiseleva@microsoft.com} \\\And
  Maarten de Rijke \\
  University of Amsterdam \\
  \href{mailto:m.derijke@uva.nl}{m.derijke@uva.nl} \\
}

\begin{document}
\maketitle
\begin{abstract}
   Automatic text summarization has experienced substantial progress in recent years. With this progress, the question has arisen whether the types of summaries that are typically generated by automatic summarization models align with users' needs. \citet{ter2020makes} answer this question negatively. Amongst others, they recommend focusing on generating summaries with more graphical elements. This is in line with what we know from the psycholinguistics literature about how humans process text. Motivated from these two angles, we propose a new task: \textit{summarization with graphical elements}, and we verify that these summaries are helpful for a critical mass of people. We collect a high quality human labeled dataset to support research into the task. We present a number of baseline methods that show that the task is interesting and challenging. Hence, with this work we hope to inspire a new line of research within the automatic summarization community.
\end{abstract}

%!TEX root = ../main.tex

\section{Introduction}
\label{sec:introduction}

Automatic text summarization has experienced substantial progress over the last couple of years, with the introduction of neural sequence to sequence models~\cite[e.g.,][]{cheng2016neural, see2017get, vaswani2017attention} and pretrained large language models~\cite[e.g.,][]{devlin-etal-2019-bert, liu2019text, lewis-etal-2020-bart}. 
There have been substantial improvements on automatic evaluation scores like ROUGE~\cite{lin2004rouge} and human evaluation metrics such as fluency.
Nevertheless, recent work has questioned whether the field is progressing in the right direction.

Specifically, in line with~\citet{jones1999automatic}, \citet{ter2020makes} have argued that the users of automatically generated summaries are often ignored when designing automatic summarization methods. By means of a survey amongst heavy users of automatically generated summaries, they show that users' needs do not fully align with current approaches to automatic text summarization. Amongst others, participants indicate being interested in summaries that contain more graphical elements, such as arrows and colored text. In this work we build upon the conclusions and recommendations of~\citet{ter2020makes} and introduce a new task: \emph{summarization with graphical elements}. 

\begin{table}
  \centering
  \begin{tabularx}{\columnwidth}{X}
  \toprule
    Laura participated in a triathlon competition. She had trained really hard and won a golden medal. The competition took place in Germany. For Laura it was her first time in Germany.  \\
    \midrule
    \includegraphics[width=0.9\columnwidth]{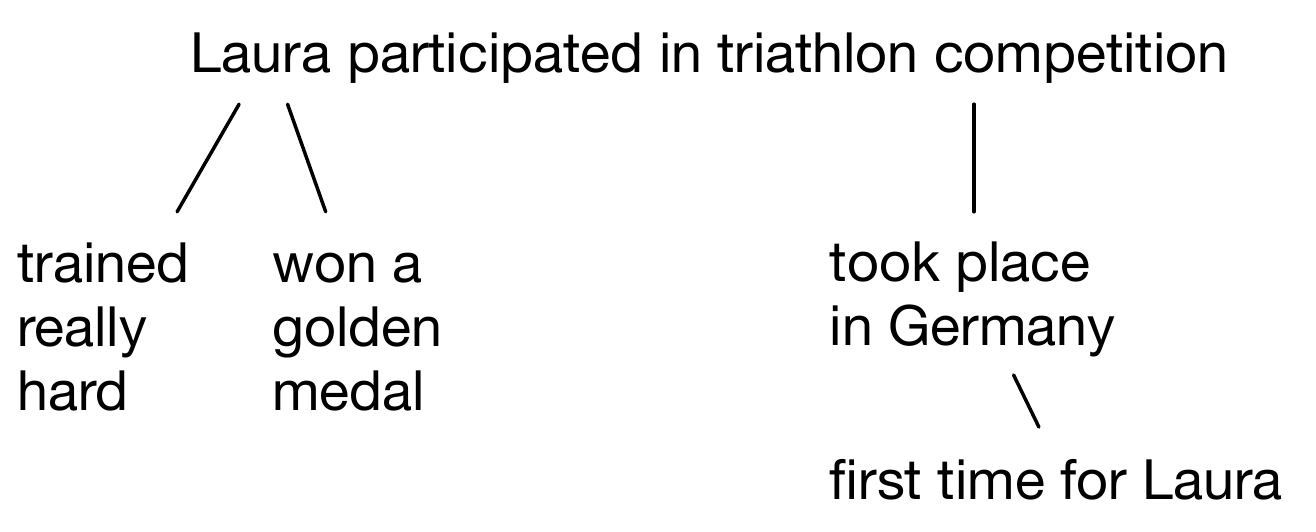} \\
  \bottomrule
  \end{tabularx}
  \caption{Example of given-new strategy of human text comprehension. Adapted based on an example in~\citet{carroll2008psychology}.}
  \label{tab:given_new_strategy}
 \end{table}
 
 \begin{figure*}
    \centering
    \includegraphics[width=\textwidth]{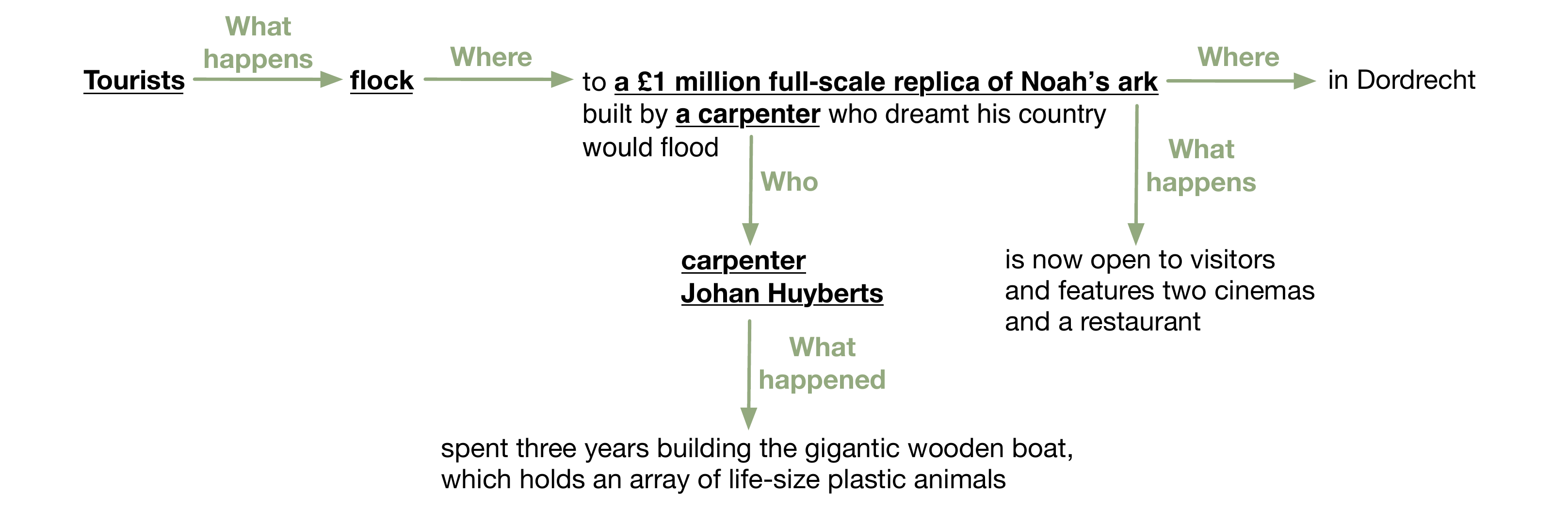}
    \caption{Example of a summary with graphical elements. In this example, nodes with outgoing edges are bold faced and underlined. The relations are marked with arrows, with the type of relation written on the arrow.}
\label{fig:example_summary_with_graphical_elements}
  \centering
\end{figure*}

In designing our task, we are also informed by the cognitive science and psycholinguistic literature on human text understanding. A popular model is the 
\textit{given-new strategy}~\cite{clark1974psychological, haviland1974s, clark1977discourse}, which states that humans attach new information to already known, i.e., \textit{given}, information in their memory, in order to build up a mental model of the information as a whole. Table~\ref{tab:given_new_strategy} shows an example. In the first row one can read a short story. The second row depicts how new information is attached to given information as one continues to read the text. While building up this mental model, humans (unconsciously) select which information to keep, and which information can be forgotten~\cite{kintsch1978toward}. That is, this process can be intuitively linked to summarization, as also noted by~\citet{cardenas2021unsupervised}.

We combine this psycholinguistic perspective with the more HCI-centered perspective from~\citet{ter2020makes} to arrive at our task of \textit{summarization with graphical elements}. The graphical elements that envisage are part of the output summaries. Figure~\ref{fig:example_summary_with_graphical_elements} shows an example of such a summary. We detail the task description in Section~\ref{sec:task_description}. We collect a high quality human labeled dataset in order to be able to evaluate model performance on the task, which we detail in Section~\ref{sec:dataset_human_labeling}. We confirm that a critical number of people are interested in these types of summaries for a variety of different tasks (Section~\ref{sec:results_human_eval_q1}).

The main contributions made in this paper are:
\begin{enumerate}[leftmargin=*,label=\textbf{C\arabic*},nosep]
    \item We introduce a new task: \emph{summarization with graphical elements};
    \item We collect a dataset, which we call \OurDataset, with high quality human labels to support research into the task; and
    \item We present the first baseline results on the task, which show that the task is challenging and can inspire a lot of future work in this direction.
\end{enumerate}

\noindent%
We make the code to run the experiments and to obtain the dataset freely available.\footnote{\url{http://www.github.com/maartjeth/summarization_with_graphical_elements}}

%!TEX root = ../main.tex

\section{Related Work}
\label{sec:related_work}

We discuss related work on automatic text summarization and on information extraction.

\subsection{Automatic text summarization}
\label{sec:related_work_automatic_text_summarization}

\paragraph{Context factors.} \citet{jones1999automatic} formulates three \textit{context factors} for automatic summarization: \begin{enumerate*}[label=(\roman*)]
    \item \textit{input}, 
    \item \textit{purpose}, and 
    \item \textit{output factors}.
\end{enumerate*} These factors describe \begin{enumerate*}[label=(\roman*)]
    \item what the input to the summarization system looks like, 
    \item what the goal of the summary is, and
    \item what the final summary looks like.
\end{enumerate*}
A large part of the research on automatic summarization focuses on generating a condensed textual version (\textit{output factors}) of a single or multiple document(s), often in the news or Wikipedia domain (\textit{input factors})~\cite[e.g.,][]{see2017get, koupaee2018wikihow, liu2019text, lewis-etal-2020-bart}. As noted by~\citet{jones1999automatic}, but also by~\citet{mani2001automatic} and by~\citet{ter2020makes}, the \textit{purpose factors} are often overlooked.

Recently, we have seen increased interest in different interpretations of the summary factors, especially in terms of the input factors. Examples include timeline summarization~\cite[e.g.,][]{li-etal-2021-timeline, yu-etal-2021-multi}, opinion summarization~\cite[e.g.,][]{angelidis-etal-2021-extractive, bravzinskas2021learning} and dialogue summarization~\cite[e.g.,][]{feigenblat-etal-2021-tweetsumm-dialog, liu-etal-2021-topic-aware}. 
A smaller body of work focuses on different interpretations of the output factors. For example, faceted summarization~\cite[e.g.,][]{meng-etal-2021-bringing} aims at constructing structured summaries. Other examples include building concept maps~\cite{falke2017bringing} and knowledge graphs~\cite{wu2020extracting}. Our work also contributes a different interpretation of the output factors, as we construct summaries with graphical elements. An important distinguishing factor from the work on concept maps and knowledge graphs is that we build upon the recommendations of~\citet{ter2020makes} and take the purpose factors into account when designing our task. That is, we explicitly focus on the usefulness of our summaries for users. As a result, our output summaries contain longer textual phrases, placing our work in between the work on concept maps and knowledge graphs and  more classic work on textual summarization.

\paragraph{Abstractive and extractive summarization.} The field of automatic summarization can classically be divided into abstractive~\cite[e.g.,][]{gehrmann-etal-2018-bottom, lewis-etal-2020-bart} and extractive summarization~\cite[e.g.,][]{narayan-etal-2018-ranking, ju-etal-2021-leveraging-information}. In our work the summaries cannot be generated by simply copying from the input document, classifying our task as a form of abstractive summarization.

\paragraph{Modeling.} Many recent automatic summarization methods rely on some variety of neural sequence to sequence modeling~\cite[e.g.,][]{cheng2016neural, vaswani2017attention, lewis-etal-2020-bart, xue-etal-2021-mt5}. Summaries produced by recent approaches such as BART~\cite{lewis-etal-2020-bart} and T5~\cite{xue-etal-2021-mt5} are of very high quality in terms of fluency and grammaticality, yet they struggle with factual consistency~\cite[e.g.,][]{cao-etal-2020-factual, maynez-etal-2020-faithfulness}. Hence, there has been a surge in work that focuses on improving the factuality of generated summaries. These works focus either on the evaluation of summarization~\cite[e.g.,][]{wang2020asking, durmus2020feqa}, or on the modeling procedures themselves, for example by explicitly incorporating graph-based meaning representations in the modeling process~\cite{ribeiro2022fact}. In this work we use BART and T5 as the summarization backbone of our baselines, but we are excited to explore graph-based methods in future work.

\subsection{Information extraction}
\label{sec:related_work_information_extraction}
The summaries in our task can be represented as relation triples. For example, \textit{(a \pounds1 million full-scale replica of Noah's ark, \textsc{where}, in Dordrecht)} would be one of the triples for the summary in Figure~\ref{fig:example_summary_with_graphical_elements}. In Section~\ref{sec:task_description} we give a detailed and formal description of our task. For our baselines we extract these summary triples from the summaries generated by BART and T5, linking our work to work on information extraction (IE). A number of neural approaches have been introduced for IE~\cite[e.g.,][]{qian2018graphie, luan2019general, wadden2019entity}. We make use of \textsc{DyGIE++}~\cite{wadden2019entity}, a BERT-based IE architecture for different IE tasks. We specifically use it for relation extraction.

%!TEX root = ../main.tex

\section{Task description}
\label{sec:task_description}
In this section we describe the summarization with graphical elements task more formally. Given an input document $D = [x_0, \ldots, x_n]$ where $x_i$ refers to the $i^{th}$ token in document $D$, our task is to generate summarizing triples of the form $(y_{a_0} \ldots\ y_{a_k}, relation, y_{b_0} \ldots\ y_{b_k})$, where $y_{a_i}$ and $y_{b_i}$ are tokens, generated in an abstractive fashion. The triples can be thought of in a more graphical way as $(y_{a_0} \ldots\ y_{a_k})$ being a node with an outgoing edge and $(y_{b_0} \ldots\ y_{b_k})$ as a node with an incoming edge. The connecting edge is labeled with \textit{relation}.

In order to improve the conciseness of the summary, our objective is to merge nodes with outgoing edges that refer to the same entity, making coreference resolution an important part of the task. This approach is in line with the given-new strategy. As an example, recall Table~\ref{tab:given_new_strategy}, where the phrases \textit{trained really hard} and \textit{won a golden medal} are linked to \textit{Laura}, instead of making a new node for \textit{She}, which is written in the original text.

The relations could be of any form, ranging from an open set to an empty set. In this work we choose to use a closed set of relations $L$. Explicitly labeling the relations, as opposed to leaving them empty, makes for a more interesting task. Given the nature of our data, in this work we define $L = \{$\emph{who}, \emph{what}, \emph{what happens}, \emph{what happened}, \emph{what will happen}, \emph{where}, \emph{when}, \emph{why}$\}$. We discuss our dataset in more detail in the next section. 

%!TEX root = ../main.tex

\section{Dataset description}
\label{sec:dataset_description}

For our task we collect a human-labeled dataset, that we call \OurDataset, short for \textit{summaries with graphical elements}. We seek to collect summary triples for each input document. Below, we explain our data selection and detail how we obtained the human labels. We end this section with a detailed description of the dataset's statistics.

\subsection{Requirements}
In choosing the specifics of \OurDataset, we need to satisfy a number of requirements. First, the domain of the data needs to be fully understandable by human annotators in order to ensure high quality annotations. That is, we cannot choose a domain that can only be fully understood by domain experts, such as scientific documents. Moreover, annotators need to be able to fluently speak the language of the data.
Secondly, the data naturally fits the task description, i.e., it is clear how summary triples can be constructed in a meaningful way.
Finally, the data needs to be easily accessible for others to reproduce and build upon our work.

\subsection{Decisions} 
\paragraph{Type of data.} Keeping the requirements in mind, we opt to use the CNN/DM dataset~\cite{hermann2015teaching} as the basis of our data collection. The dataset fits all listed requirements: \begin{enumerate*}[label=(\roman*)]
\item news documents do not require specific domain knowledge from annotators and our annotators are fluent in English, the language of the dataset; 
\item the story highlights from the CNN/DM dataset give us a way to construct abstractive summary triples and we can use the set of $5$W's --- that are in the nature of news articles and have been used for automatic summarization before~\cite{parton2009comparing} --- as our set of relations; and
\item we are able to release the code to obtain a labeled version of the dataset, so that our work is reproducible.
\end{enumerate*}
We acknowledge that the CNN/DM dataset and the (English) news domain in general have been studied extensively, yet a thorough exploration of alternatives convinced us that the CNN/DM dataset fits our requirements best to start with. We hope that more datasets will be collected for summarization with graphical elements in different domains and languages in the future.

As mentioned, we choose to use the $5$W's (\textit{who, what, where, when, why}) as our relations and we choose to add three additional labels to be able to provide more temporal nuance and make the task more challenging: \textit{what happens, what happened} and \textit{what will happen}.

\paragraph{Including the title.} In preparing our labeling task, we noticed that many of the story highlights are hard to understand without the title of the news article, as the highlights often refer back to information that was introduced in the title. Therefore, we add the titles to the summary abstracts. In our labeling procedure we confirm our intuition that the title is essential in $80\%$ of the abstracts (Section~\ref{sec:dataset_statistics}).

\paragraph{Scale of human labeling.} Given the intensity of the labeling procedure (see below), we opt to collect a human-labeled \textit{test set}. Each document is labeled by three annotators. This allows us to account for the ambiguity in the summarization process during evaluation, i.e., there are multiple ways of correctly constructing a summary for a single document. We use this opportunity to shuffle the standard train, validation and test sets of the CNN/DM dataset. From the entire set, we select $500$ documents for human labeling, which we publish with our code. The remaining documents can be used for a train and validation set.

\subsection{Human labeling}
\label{sec:dataset_human_labeling}

In order to construct a high quality test set, we recruit three annotators with NLP expertise. The annotators need to construct summary triples, based on input abstracts from the CNN/DM dataset.
Annotators are instructed via a detailed instruction manual, a video call in which the manual was discussed and there was room for communication via a chat channel. This allowed annotators to ask questions while doing the annotations, but also to report mistakes, which improved the quality control of the annotations (Section~\ref{sec:dataset_quality_control}). Annotators were paid an hourly rate, removing the incentive to rush responses. Annotators were first asked to annotate a set of six articles, after which they received feedback on their annotations. During the remaining annotation task we checked the quality of the incoming annotations, by sampling annotations at random and inspecting them manually, to make sure annotators were still on the right track. All annotators annotated the same set of documents, resulting in three annotations per document (apart from a handful of documents where something went wrong with the annotations, see Section~\ref{sec:dataset_quality_control}). 

Each document is annotated in a Human Intelligence Task (HIT). 
Examples are given in Appendix~\ref{sec:appendix_human_labeling}. Each HIT consists of the following:

\begin{enumerate}[leftmargin=*,nosep,label=(\arabic*)]
  \item \textit{Introduction.} Here we iterate the most important parts of the instruction manual.
  \item \textit{The actual task.} Annotators are presented with an abstract taken from our test set, including the title. The abstract is explicitly divided into sentences. We automatically divide the abstract into constituents with the Berkely constituency parser~\cite{kitaev2018constituency}, again per sentence. Annotators then need to select the relation triples per the guidance in the annotation manual. Annotators have the option to select that something was wrong with the presented abstract or the constituents. Annotators can also indicate if the first sentence of the abstract (i.e., the title of the article) was fully redundant. If they click that option, we ask them to not select any constituents from the first sentence. Lastly, annotators can check a box if they were particularly uncertain about their annotations. 
\end{enumerate}

\subsubsection{Quality control}
\label{sec:dataset_quality_control}

 \begin{table}
  \centering
  \begin{tabular}{lr}
  \toprule
    \textbf{Metric} & \textbf{Avg} $\pm$ \textbf{Std} \\
    \midrule
    Hard F1 & $0.21 \pm 0.16$ \\
    Soft F1 & $0.47 \pm 0.18$ \\
    Jaccard w.r.t. Triple & $0.13 \pm 0.11$ \\
    Jaccard w.r.t. Const A & $0.28 \pm 0.16$ \\
    Jaccard w.r.t. Const B & $0.32 \pm 0.16$ \\ 
    Jaccard w.r.t. Relations & $0.61 \pm 0.17$ \\ 
  \bottomrule
  \end{tabular}
  \caption{Overlap statistics of annotators on human labeled test set. \textit{Const A} refers to the first and \textit{Const B} to the second constituent in a summary triple.}
  \label{tab:agreement_human_annotators}
  \end{table}

Within our budget we were able to obtain annotations for $295$ documents. Annotators spent on average around $7$ minutes on each HIT. In addition to the quality control we did during the annotation process, we also inspect annotators' answers to the quality control questions: \begin{enumerate*}[label=(\roman*)]
    \item whether something was wrong with the abstract or presented constituents, and
    \item whether they indicated to be uncertain about their annotations. We also analyse the overlap in annotations per HIT. 
\end{enumerate*}

\paragraph{Issues with the presented abstract or constituents.} Here we manually inspect the annotations of the HITs where a majority of the annotators indicated that there was something wrong, such as a mistake with the automatically generated constituents. This does not necessarily mean that the annotation is also wrong. We discard eight documents based on these answers. Moreover, we discard one more article because of preprocessing issues with the abstract later in the pipeline.

\paragraph{Annotator uncertainty.}
We manually inspect all annotations where annotators indicated to be uncertain about a HIT and we discard twelve HITs. 

\paragraph{Reported issues by annotators.} Annotators had the option to report mistakes they made via chat. For example, sometimes annotators realised after submitting a HIT that they chose an incorrect label. Based on these reports we manually made changes to the annotations of six documents.

\paragraph{Overlap in annotations.} For each HIT, we compute the pairwise macro $F_1$-scores for all annotator pairs. We compute both hard and soft scores. For the hard scores, we only count a selected triple if both annotators have exactly that triple in their annotations.
However, this type of scoring is extremely conservative, especially given the nature and the ambiguity of the annotation task. Intuitively, we also want to assign points if the triple partially overlaps, but not entirely, for example because one annotator decided to include an article, whereas another annotator did not. Therefore, we also compute a soft score, where we greedily align the best matching triples and compute the lexical overlap between each of these and average them. We also compute the average Jaccard scores for annotators pairs. Specifically, we compute these scores for the entire triple, and for each individual component of the summary triple. These statistics are given in Table~\ref{tab:agreement_human_annotators}. From these scores, it becomes clear that the annotators are aligned in their annotations, yet it also shows that there are different ways to construct the summaries. This underlines our choice of collecting multiple annotations per data point. 
During evaluation of the task, the best matching annotation can be chosen as ground truth.

\subsection{Dataset statistics}
\label{sec:dataset_statistics}
We present the statistics for the collected test set in Table~\ref{tab:dataset_statistics}. After quality control, our dataset consists of $286$ documents, which is comparable to earlier work that constructed human annotated test sets for summarization~\cite[e.g.,][]{wu2020extracting}. The vast majority of documents have three annotations. We also confirm our intuition about including the titles: in almost $80\%$ of the cases the title was needed to understand the summary abstract. Lastly, we find that all relations are represented, with some very popular relations such as \textit{what happened} and \textit{what}.

\begin{table}
  \centering
  \begin{tabular}{lrr}
  \toprule
  & \textbf{Counts} & \textbf{\%} \\
  \midrule
  $\#$ Documents & $286$ & $100$ \\
  \midrule
  $\#$ Three annotations &  & \\
  per document & $268$  & $93.7$ \\
  $\#$ Two annotations & & \\
  per document & $18$ & $6.29$ \\
  \midrule
  $\#$ Triples & $5942$ & $-$ \\
  Avg $\#$ triples  &  & \\
  per document & $7.07 \pm 3.28$ & $-$ \\
  \midrule
  Title redundant &  & \\
  (majority vote) & $58$ & $20.3$ \\
  \midrule
  $\#$ Who & $439$ & $7.39$ \\
  $\#$ What & $1,407$ & $23.7$ \\
  $\#$ What happens & $967$ & $16.3$ \\
  $\#$ What happened & $2,075$ & $34.9$\\
  $\#$ What will happen & $149$ & $2.51$\\
  $\#$ Where & $339$ & $5.71$ \\
  $\#$ When & $333$ & $5.60$ \\
  $\#$ Why & $233$ & $3.92$ \\
  \bottomrule
  \end{tabular}
  \caption{Dataset statistics of \OurDataset.}
  \label{tab:dataset_statistics}
  \vspace{-2pt}
  \end{table}

%!TEX root = ../main.tex

\section{Baselines}
\label{sec:method}

We provide a number of baselines for our task. All baselines consist of two steps: \begin{enumerate*}[label=(\roman*)]
    \item a summarization step, and
    \item a relation extraction step.
\end{enumerate*} We leave entire end-to-end solutions as an interesting and important direction for future work. For each of the steps, we choose well-performing and easily-accessible methods, that we adapt for our task where needed. We discuss each step below.

\subsection{Step 1 -- Summarization}
For the summarization step, we finetune BART-large~\cite{lewis-etal-2020-bart} and T5~\cite{raffel2019exploring}. We train and validate the models on our train and validation splits for the CNN/DM dataset (Section~\ref{sec:dataset_description}). We report the inference scores on the abstracts in our test set. We use the Huggingface library\footnote{\url{https://github.com/huggingface/transformers}}~\cite{wolf-etal-2020-transformers} and we finetune both models on $4$ GPUs with $12$GB of RAM each.

\subsection{Step 2 -- Relation extraction}

We investigate two approaches for the relation extraction, which we discuss below.

\subsubsection{Creating labels with Snorkel}
We make use of Snorkel~\cite{ratner2017snorkel} as a means to obtain summary triples. Snorkel is a well-documented method\footnote{\url{https://www.snorkel.org/}} that generates (weak) labels for unlabeled data using on (noisy) labeling functions and trained generative models on top of these labeling functions. We use the obtained summary triples for two purposes: \begin{enumerate*}[label=(\roman*)]
    \item to extract relations directly on the output of the summarization models, and
    \item to use as weak labels for training a relation extraction model.
\end{enumerate*} Our Snorkel pipeline consists of three stages: \begin{enumerate*}[label=(\roman*)]
\item candidate pairs selection,
\item relation labeling, and
\item final filtering
\end{enumerate*}. We discuss these stages in more detail below. 

\paragraph{Candidate pairs selection.}
We use the Berkeley constituency parser~\cite{kitaev2018constituency} as a fast and high quality parser to obtain all constituents in a summary abstract. Next, we combine the constituents to make potential candidate pairs. Naively, one could simply combine all constituents that are found for an abstract. However, due to its quadratic complexity, this approach is very resource demanding. Instead, we make use of a few heuristics to make the candidate pairs. First, we discard all single token constituents that are of type \textsc{IN}, \textsc{DT} or \textsc{CC}, as these would not lead to valid candidate pairs. For the same reason, we also discard all constituents that only consist of a special token, such as a punctuation mark. Moreover, we make sure to not pair overlapping constituents. Finally, we set a threshold that constituents pairs can be at most two sentences apart. With this heuristic, we miss a small fraction of correct candidate pairs, yet it considerably speeds up the computation.

\paragraph{Relation labeling.}
In the second step, we construct labels for relations between the candidate pairs. For each possible relation, we construct Snorkel labeling functions. These functions are fuzzy heuristics, that together form a basis to train a model that gives a final weak label for a data point. In our work, we use these weak labels directly, and we also use them to train a relation extraction model (Section~\ref{sec:method_dygiepp}). 

\paragraph{Final filtering.}
As a last step, we apply filtering to obtain the final set of weakly labeled data points. First, we determine the edge directions. Let us take constituents connected by the \textsc{when} relation as an example. In these cases, the constituent that indicates the time should be the second constituent in the summary triple. Next, we filter out overlapping constituents. For example, imagine three possible triples: \textit{(event, \textsc{when}, June 7)}, \textit{(event, \textsc{when}, 2012)}, \textit{(event, \textsc{when}, June 7, 2012)}. In this case, we make sure that only one of these potential data points is included in the weakly labeled training set. As a heuristic, we choose to include the data point with the longest string, arguing that this provides us with most information. Finally, we merge all coreferences and use the first occurrence of the referent in our summary triples, corresponding to the instructions for our human annotators. We use AllenNLP's\footnote{\url{https://allenai.org/allennlp}} implementation of SpanBERT\footnote{\url{https://github.com/allenai/allennlp-models/blob/main/allennlp_models/modelcards/coref-spanbert.json}}~\cite{joshi-etal-2020-spanbert} as a coreference parser, as we found that this parser gave us the most reliable results.

\subsubsection{Trained relation extraction}
\label{sec:method_dygiepp}
We choose \textsc{DyGIE++}~\cite{wadden2019entity} as our trained relation extraction method. \textsc{DyGIE++} has  shown to be a well-performing method across several tasks and datasets, it is well documented and can be easily adapted to our task. We use the code as provided,\footnote{\url{https://github.com/dwadden/dygiepp}} but make a number of adaptations. 

First, the standard way of training \textsc{DyGIE++} is to select relations on a per sentence basis, whereas our relations span multiple sentences. As mentioned in the original work, \textsc{DyGIE++} can also be used to select relations on the document level, i.e., cross-sentence. For our implementation, this means that we do not split the summaries in sentences, but treat the entire document as if it were a single sentence. We leave all punctuation marks in. We train \textsc{DyGIE++} on $4$ GPUs with $12$GB RAM each. A cross-sentence approach is substantially more memory demanding than a within sentence approach. For this reason, we resort to training \textsc{DyGIE++} on $10,000$ summary abstracts from the training set, the maximum number of abstracts we could process with our machines.\footnote{We also found that results did not substantially improve by adding more training data.} As a second adaptation to the original work, we evaluate the extracted relations on the same metrics that we evaluated the human annotations on.
%!TEX root = ../main.tex

\section{Results}
\label{sec:results}

We evaluate our work from three angles, which we discuss in this section. The first question we ask is whether a critical mass of people is interested in our suggested summaries with graphical elements. We confirm that this is the case and present our setup for a human evaluation to test this in Section~\ref{sec:results_human_eval_q1}. Secondly, we evaluate our baseline models. We present an automatic evaluation in Section~\ref{sec:results_baselines_automatic} and a human evaluation in Section~\ref{sec:results_baselines_human}. Thirdly, we perform a qualitative analysis of the performance of the baseline methods in Section~\ref{sec:results_qualitative}. 

\subsection{Q1 -- Human evaluation}
\label{sec:results_human_eval_q1}

First, we evaluate users' preference for different types of summaries in different settings and scenarios. Specifically, we want to find out whether a critical mass of people is interested in the summaries with graphical elements. We stress that we do not necessarily aim for a majority of people wanting to use a summary with graphical elements. As noted in previous work~\cite{jones1999automatic, ter2020makes}, different people have different preferences in different contexts. For our evaluation we use the recommendations from~\citet{ter2020makes}. That is, we make use of a simulated work task setting~\cite{borlund2003iir} to evaluate three types of summaries in a pairwise manner on two purpose factors. Our scenario outline is the following: \textit{Imagine that you would like to quickly gather information about a certain news event. To help you quickly find your information, you have access to a summary that describes the news article.}

The three summaries that we compare are: 
\begin{enumerate}[leftmargin=*,nosep, label=(\arabic*)]
    \item \textit{A text only summary.} For this we simply use the summary abstracts.
    \item \textit{A summary with graphical elements.} For this we use a human labeled ground truth summary. We manually convert the labeled triples to a graphical summary. We acknowledge that the precise lay-out may affect people's judgements. The results of this human evaluation should therefore be taken as an indication of people's preference and taking this work in production should include an additional design step. 
    \item \textit{Typeset control summary.} To control for some of the bias discussed in the previous point, we add a summary that is pure text, yet contains text formatting: we bold face the first sentence and color all phrases that contain outgoing edges in the human labeled equivalent.
\end{enumerate}

\noindent%
We evaluate the summaries on the purpose factors \textit{previewing} and \textit{substituting} as defined in~\citet{jones1999automatic} and used in~\citet{ter2020makes}. That is, we ask the following two questions: \textit{Which summary is more useful to you to preview the full news article?} and \textit{Which summary is more useful to you to substitute the full news article?}

We run this evaluation on Amazon Mechanical Turk.\footnote{\url{http://www.requester.mturk.com/}} We compare summaries for two different news articles, to control for potential bias towards an article. We randomize the position of the articles on the page, to avoid position bias. We request $20$ judgements per comparison per news article, i.e., $40$ pairwise judgements in total. The crowd workers are U.S. based, have a HIT approval rate of at least $90\%$ and at least $1000$ accepted HITs. 

For quality control, we add three questions at the start of each HIT. The first two questions are about the content of the summary and are different for each summary pair. We manually construct these questions based on the ground truth. We list the questions in Appendix~\ref{sec:open_questions_human_eval_q1}. In the third question, we ask which summary workers mainly used to answer the question. Significant failure to answer the first two questions results in a rejection of the HIT and we request new labels for the rejected HITs. 
In this manner we arrive at a total of $120$ judgements. A full overview of the task is given in Figure~\ref{fig:human_eval_q1_overview} in Appendix~\ref{sec:appendix_human_eval_q1}.

\paragraph{Findings.} In the main body of the paper, we report the aggregated results on both news articles in Table~\ref{tab:human_eval_q1_all}, as there are no substantial differences if we split the results per document. The results per document are given in Appendix~\ref{sec:appendix_human_eval_q1}, in Table~\ref{tab:human_eval_q1_doc1} and~\ref{tab:human_eval_q1_doc2}. From these results there are two main observations. First, a substantial fraction of crowd workers prefer a summary with graphical elements over the other two summaries for both purpose factors and a substantial fraction of the crowd workers used the graphical summary to find the answer to the questions about the content of the summary. 
It is not a majority, yet the group is clearly large enough to be convinced that summaries with graphical elements are an important focus point, which is in line with \citet{ter2020makes}. Second, the typeset summaries are very popular, much more so than the raw text summaries. This, again, points in the direction that adding graphical elements to summaries is an important research direction. 

\begin{table}

\begin{tabular}{llrr}
\toprule
	& \multicolumn{1}{c}{\textbf{Pair (A/B)}} & \multicolumn{1}{c}{\textbf{Prefer A}} & \multicolumn{1}{c}{\textbf{Prefer B}} \\
	& \multicolumn{1}{c}{} & \multicolumn{1}{c}{\textbf{(\%)}} & \multicolumn{1}{c}{\textbf{(\%)}}\\
	\midrule
\parbox[t]{2mm}{\multirow{3}{*}{\rotatebox[origin=c]{90}{\emph{Used}}}}
        &  Graphical / Text  & $35.0$ & $65.0$  \\
        &  Graphical / Typeset  & $50.0$ & $50.0$  \\
	      &  Typeset / Text  & $72.5$ & $27.5$ \\ \hline
\parbox[t]{2mm}{\multirow{3}{*}{\rotatebox[origin=c]{90}{\emph{Prev}}}}
	&  Graphical / Text   & $35.0$  & $65.0$ \\
	&  Graphical / Typeset  & $47.5$ & $52.5$  \\
	&  Typeset / Text   & $75.0$ & $25.0$ \\ \hline
\parbox[t]{2mm}{\multirow{3}{*}{\rotatebox[origin=c]{90}{\emph{Sub}}}}
	&  Graphical / Text  & $30.0$ & $70.0$ \\
	&  Graphical / Typeset  &$32.5$  & $67.5$ \\
	&  Typeset / Text  & $65.0$ & $35.0$  \\
	\bottomrule
\end{tabular}
\caption{Results Human Evaluation Q1. Pairwise comparisons.}
\label{tab:human_eval_q1_all}
\end{table}

\subsection{Q2 -- Automatic evaluation}
\label{sec:results_baselines_automatic}

The second question that we evaluate is which baseline model performs best on our newly proposed task. 
First, we confirm that our summarization models score on par with the scores that are reported when evaluating on the entire test set from the CNN/DM dataset (Table~\ref{tab:results_automatic_eval_summarization}).

Next, we evaluate the different relation extraction components. Apart from the summarization models with Snorkel and \textsc{DyGIE++} labels, we also include the scores of applying Snorkel and \textsc{DyGIE++} to the ground truth CNN/DM abstracts directly. We evaluate according to the same metrics as for the human labels and we use the best scoring ground truth to compute the final scores.\footnote{Some documents could not be parsed in the Snorkel pipeline and we leave these out.} 

\paragraph{Findings.} The results are given in Table~\ref{tab:results_automatic_eval_pipeline_hard}. We also report additional metrics in Appendix~\ref{sec:appendix_automatic_eval} in Table~\ref{tab:results_automatic_eval_pipeline_soft}. From the results a few things become clear. In general, we find that the scores are still far from the agreement scores of the human labeled annotations. This indicates that this is an interesting task with a lot of opportunity for future work. More specifically, we see that applying Snorkel labels directly gives us better results than training \textsc{DyGIE++} on these labels. The additional summarization step also decreases the performance substantially, even though the produced summaries by both BART and T5 seem to be of high quality. We postulate that these summaries are still quite different from the human-written summaries, therefore decreasing the performance of the models that were trained on the human-written summaries.

We also compare the relations that are predicted in different setups. We find that the types of relations that are predicted are in line with the scores we find for the automatic metrics. Methods with lower recall scores predict fewer relations. In general, the distribution of relation counts are comparable amongst methods, but differ more for methods with lower scores. We share histograms in Appendix~\ref{sec:appendix_automatic_eval}, Figure~\ref{fig:hists_relations_counts_baselines} and~\ref{fig:hists_relations_percentage_baselines}.

\begin{table}
\centering
\begin{tabular}{lrrrr}
\toprule
\textbf{Model} & \textbf{R1} & \textbf{R2} & \textbf{RL} & \textbf{RLsum} \\
\midrule
BART-L & $48.16$ & $22.81$ & $32.84$ & $44.49$ \\
T5 & $46.63$ & $22.41$ & $32.97$ & $43.43$ \\
\bottomrule
\end{tabular}
\caption{Rouge scores summarization component.}
\label{tab:results_automatic_eval_summarization}
\end{table}

\begin{table*}
\centering
\begin{tabular}{l rrr r}
\toprule
 & \multicolumn{3}{c}{\textbf{Hard}}  & \multicolumn{1}{c}{\textbf{Soft Greedy}}    \\
\cmidrule(lr){2-4} \cmidrule(lr){5-5} 
&  \multicolumn{1}{c}{\textbf{P}} & \multicolumn{1}{c}{\textbf{R}} & \multicolumn{1}{c}{\textbf{F1}} & \multicolumn{1}{c}{\textbf{F1}}  \\
\midrule
GT Abstract + Snorkel & $0.130 \pm 0.165$ & $0.102 \pm 0.136$ & $0.111 \pm 0.141$ & $0.446 \pm 0.121$ \\
GT Abstract + \textsc{DyGIE++} &  $0.071 \pm 0.157$ & $0.037 \pm 0.084$ & $0.046 \pm 0.098$ & $0.256 \pm 0.145$ \\
\midrule
BART + Snorkel & $0.003 \pm 0.024 $ & $ 0.002 \pm 0.013$ & $0.002 \pm 0.017$  & $0.323 \pm 0.087$ \\
T5 + Snorkel & $0.008 \pm 0.056 $ & $0.004 \pm 0.027$ & $0.005 \pm 0.035$ & $ 0.329 \pm 0.086$  \\
\midrule
BART + \textsc{DyGIE++} & $0.001 \pm 0.015$ & $0.00 \pm 0.007$ & $0.001 \pm 0.009$  & $0.188 \pm 0.120$ \\
T5 + \textsc{DyGIE++} & $0.003 \pm 0.033$ & $0.001 \pm 0.009$ & $0.001 \pm 0.014$ & $0.184 \pm 0.115$  \\
\bottomrule
\end{tabular}
\caption{Results for different methods on our task of generating summaries with graphical elements.}
\label{tab:results_automatic_eval_pipeline_hard}
\vspace{2pt}
\end{table*}

\subsection{Q2 -- Human evaluation }
\label{sec:results_baselines_human}

\begin{table}[h!]
\vspace{5pt}
\begin{tabular}{llrr}
\toprule
	& \multicolumn{1}{c}{\textbf{Pair (A/B)}} & \multicolumn{1}{c}{\textbf{Prefer A}} & \multicolumn{1}{c}{\textbf{Prefer B}} \\
	& \multicolumn{1}{c}{} & \multicolumn{1}{c}{\textbf{(\%)}} & \multicolumn{1}{c}{\textbf{(\%)}}\\
	\midrule
\parbox[t]{2mm}{\multirow{3}{*}{\rotatebox[origin=c]{90}{\emph{Inform}}}}
        &  Human / Snorkel  & $80.0$ & $20.0$  \\
        &  Human / \textsc{DyGIE++}  & $93.3$ & $6.70$  \\
	    &  Snorkel / \textsc{DyGIE++}  & $86.7$ & $13.3$ \\ \hline
\parbox[t]{2mm}{\multirow{3}{*}{\rotatebox[origin=c]{90}{\emph{Cons}}}}
        &  Human / Snorkel  & $60.0$ & $40.0$  \\
        &  Human / \textsc{DyGIE++}  & $73.3$ & $26.7$  \\
	    &  Snorkel / \textsc{DyGIE++}  & $73.3$ & $26.7$ \\ \hline
\parbox[t]{2mm}{\multirow{3}{*}{\rotatebox[origin=c]{90}{\emph{Prev}}}}
        &  Human / Snorkel  & $73.3$ & $26.7$  \\
        &  Human / \textsc{DyGIE++}  & $86.7$ & $13.3$  \\
	    &  Snorkel / \textsc{DyGIE++}  & $80.0$ & $20.0$ \\ \hline
\parbox[t]{2mm}{\multirow{3}{*}{\rotatebox[origin=c]{90}{\emph{Sub}}}}
        &  Human / Snorkel  & $86.7$ & $13.3$  \\
        &  Human / \textsc{DyGIE++}  & $100.0$ & $0.00$  \\
	    &  Snorkel / \textsc{DyGIE++}  & $13.3$ & $86.7$ \\ \hline
\parbox[t]{2mm}{\multirow{3}{*}{\rotatebox[origin=c]{90}{\emph{Fav}}}}
        &  Human / Snorkel  & $86.7$ & $13.3$  \\
        &  Human / \textsc{DyGIE++}  & $93.3$ & $6.7$  \\
	    &  Snorkel / \textsc{DyGIE++}  & $86.7$  & $13.3$  \\ 
	\bottomrule
\end{tabular}
\caption{Results Human Evaluation Q2. Pairwise comparisons. Results based on majority votes.}
\label{tab:results_human_eval_q2}
\end{table}

\begin{table}[h!]
\begin{tabular}{lrrr}
\toprule
	\multicolumn{1}{c}{\textbf{Pair (A/B)}} & \multicolumn{1}{c}{\textbf{Use A}} & \multicolumn{1}{c}{\textbf{Use B}} & \multicolumn{1}{c}{\textbf{No ans}}\\
	\multicolumn{1}{c}{} & \multicolumn{1}{c}{\textbf{(\%)}} & \multicolumn{1}{c}{\textbf{(\%)}} & \multicolumn{1}{c}{\textbf{(\%)}}\\
	\midrule
        Human / Snorkel  & $73.3$ & $20.0 $  & $6.7$ \\
        Human / \textsc{DyGIE++}  & $73.3$ & $13.3$ & $13.3$ \\
	    Snorkel / \textsc{DyGIE++}  & $26.7$ & $20.0$ & $53.3$ \\ 
	\bottomrule
\end{tabular}
\caption{Results Human Evaluation Q2. Which summary was used to answer the questions. Pairwise comparisons. Aggregated scores. Results based on majority votes.}
\label{tab:results_human_eval_q2_use}
\end{table}

We also evaluate the output of our baseline models with a human evaluation. Our setup is similar to the one in Section~\ref{sec:results_human_eval_q1}. Again, we compare summaries in a pairwise manner. This time, the summaries are produced by three different methods: \begin{enumerate*}[label=(\roman*)]
    \item the human labeled ground truth, 
    \item BART output followed by Snorkel labels, and
    \item BART output followed by \textsc{DieGIE++} labels.
\end{enumerate*}
We compare summaries for $15$ different articles and we request $3$ annotations per HIT to be able to compute the majority vote afterwards. We select our summaries randomly, but filter out summaries with potentially sensitive topics for crowd workers. We randomly select one of our human annotators per human-labeled summary. 

As in Section~\ref{sec:results_human_eval_q1}, we manually create summaries based on the model output. Examples are given in Appendix~\ref{sec:appendix_human_eval_q2}. We again ask questions about the content of the summary for quality control, that we construct based on the ground truth abstracts. Since not all generated summaries contain the answer to this question, we ask workers to answer `No Answer' if the summary does not contain the answer to the question. This time, we ask which summary workers used after each open question, as sometimes one summary contains the answer to one question, while the other summary the answer to the other question. We also add an additional option where workers can indicate that none of the summaries contained the answer to the question. We publish a list of selected summaries and the corresponding questions together with our code.

In addition to the questions we asked in the first round of the human evaluation, we also ask which summaries workers find more \textit{informative} and which ones they find more \textit{concise}, in line with previous work~\cite[e.g.,][]{paulus2017deep, narayan2018don}. We do not evaluate on fluency, as the nature of our summaries with graphical elements does not align with this metric. As an additional quality control question, we ask workers to indicate their favorite summary and to provide a short justification for their choice. 
As in Section~\ref{sec:results_human_eval_q1}, we obtain new annotations to replace rejected HITs and apply some filtering afterwards to ensure good quality annotations. This leaves us with a total of $134$ annotations. An example of our task is given in Appendix~\ref{sec:appendix_human_eval_q2} in Figure~\ref{fig:human_eval_q2_overview}.

\paragraph{Findings.} The results, based on majority votes, are given in Table~\ref{tab:results_human_eval_q2} and Table~\ref{tab:results_human_eval_q2_use}. These results show that crowd workers prefer the human-annotated summaries on all metrics, followed by the version where we used BART and Snorkel. Summaries with graphical elements  produced by BART and \textsc{DyGIE++} are preferred least. Table~\ref{tab:results_human_eval_q2_use} shows that these summaries often do not contain the answer to questions. These results are in line with the results on the automatic metrics.

\subsection{Q3 -- Qualitative observations}
\label{sec:results_qualitative}

We manually inspect the outputs of the different methods that we compare in the human evaluation. First, we note that \textsc{DyGIE++} still misses many relations. This is in line with the automatic scores for recall and the scores for the human evaluation. An example is given in Appendix~\ref{sec:appendix_human_eval_q2} (Figure~\ref{fig:human_eval_q2_example_1_dygiepp}). As a second observation, we note that many coreferences are missed, resulting in summaries that are less well-connected than their human-labeled counterparts. An example of this is given in Figure~\ref{fig:human_eval_q2_example_2_snorkel} in the appendix.

%!TEX root = ../main.tex

\section{Discussion}
\label{sec:discussion}

In the previous section we showed that a substantial number of users is interested in summaries with graphical elements, in line with observations from~\citet{ter2020makes}. We have also shown that current summarization and relation extraction methods still have difficulties with our task. For future work we plan to explore graph-based summarization methods to directly learn the summarization triples. Moreover, we see many opportunities to investigate model understanding with our task. As mentioned in Section~\ref{sec:related_work_automatic_text_summarization}, current automatic summarization models have difficulties with factual consistency and being able to generate correct summary triples, including correct relations, may require an additional level of factual consistency.

%!TEX root = ../main.tex

\section{Conclusion}
\label{sec:conclusion}

In this work we have proposed a new task: \textit{summarization with graphical elements}. We have collected a high quality human-labeled dataset, \OurDataset, for the task and presented the results for a number of baselines. By means of automatic and human evaluations, we show that our task is a much wanted, and challenging, addition to the existing types of automatic summarization tasks. As such, our work can inspire a lot of follow up work in this direction. For future work we are interested in learning end-to-end methods for our task and using \OurDataset~to explicitly work on factual consistency of automatic summarization models.

\section*{Acknowledgements}
We thank Evangelos Kanoulas, Nikos Voskarides, Samarth Bhargav, Dan Li and Ana Lucic for helpful comments and feedback.
This research was supported by the Nationale Politie.
All content represents the opinion of the authors, which is not necessarily shared or endorsed by their respective employers and/or sponsors.

% Entries for the entire Anthology, followed by custom entries
\bibliography{custom,anthology}

\begin{thebibliography}{48}
\expandafter\ifx\csname natexlab\endcsname\relax\def\natexlab#1{#1}\fi

\bibitem[{Angelidis et~al.(2021)Angelidis, Amplayo, Suhara, Wang, and
  Lapata}]{angelidis-etal-2021-extractive}
Stefanos Angelidis, Reinald~Kim Amplayo, Yoshihiko Suhara, Xiaolan Wang, and
  Mirella Lapata. 2021.
\newblock \href {https://doi.org/10.1162/tacl_a_00366} {Extractive opinion
  summarization in quantized transformer spaces}.
\newblock \emph{Transactions of the Association for Computational Linguistics},
  9:277--293.

\bibitem[{Borlund(2003)}]{borlund2003iir}
Pia Borlund. 2003.
\newblock The {IIR} evaluation model: a framework for evaluation of interactive
  information retrieval systems.
\newblock \emph{Information research}, 8(3):8--3.

\bibitem[{Bra{\v{z}}inskas et~al.(2021)Bra{\v{z}}inskas, Lapata, and
  Titov}]{bravzinskas2021learning}
Arthur Bra{\v{z}}inskas, Mirella Lapata, and Ivan Titov. 2021.
\newblock Learning opinion summarizers by selecting informative reviews.
\newblock \emph{arXiv preprint arXiv:2109.04325}.

\bibitem[{Cao et~al.(2020)Cao, Dong, Wu, and Cheung}]{cao-etal-2020-factual}
Meng Cao, Yue Dong, Jiapeng Wu, and Jackie Chi~Kit Cheung. 2020.
\newblock \href {https://doi.org/10.18653/v1/2020.emnlp-main.506} {Factual
  error correction for abstractive summarization models}.
\newblock In \emph{Proceedings of the 2020 Conference on Empirical Methods in
  Natural Language Processing (EMNLP)}, pages 6251--6258, Online. Association
  for Computational Linguistics.

\bibitem[{Cardenas et~al.(2021)Cardenas, Galle, and
  Cohen}]{cardenas2021unsupervised}
Ronald Cardenas, Matthias Galle, and Shay~B Cohen. 2021.
\newblock Unsupervised extractive summarization by human memory simulation.
\newblock \emph{arXiv preprint arXiv:2104.08392}.

\bibitem[{Carroll(2008)}]{carroll2008psychology}
David~W Carroll. 2008.
\newblock \emph{Psychology of language, Fifth Edition}.
\newblock Thomson Brooks.

\bibitem[{Cheng and Lapata(2016)}]{cheng2016neural}
Jianpeng Cheng and Mirella Lapata. 2016.
\newblock Neural summarization by extracting sentences and words.
\newblock \emph{arXiv preprint arXiv:1603.07252}.

\bibitem[{Clark et~al.(1977)Clark, Haviland, and Freedle}]{clark1977discourse}
Herbert~H Clark, S~Haviland, and Roy~O Freedle. 1977.
\newblock Discourse production and comprehension.

\bibitem[{Clark and Haviland(1974)}]{clark1974psychological}
Herbert~H Clark and Susan~E Haviland. 1974.
\newblock Psychological processes as linguistic explanation.
\newblock \emph{Explaining linguistic phenomena}, pages 91--124.

\bibitem[{Devlin et~al.(2019)Devlin, Chang, Lee, and
  Toutanova}]{devlin-etal-2019-bert}
Jacob Devlin, Ming-Wei Chang, Kenton Lee, and Kristina Toutanova. 2019.
\newblock \href {https://doi.org/10.18653/v1/N19-1423} {{BERT}: Pre-training of
  deep bidirectional transformers for language understanding}.
\newblock In \emph{Proceedings of the 2019 Conference of the North {A}merican
  Chapter of the Association for Computational Linguistics: Human Language
  Technologies, Volume 1 (Long and Short Papers)}, pages 4171--4186,
  Minneapolis, Minnesota. Association for Computational Linguistics.

\bibitem[{Durmus et~al.(2020)Durmus, He, and Diab}]{durmus2020feqa}
Esin Durmus, He~He, and Mona Diab. 2020.
\newblock Feqa: A question answering evaluation framework for faithfulness
  assessment in abstractive summarization.
\newblock \emph{arXiv preprint arXiv:2005.03754}.

\bibitem[{Falke and Gurevych(2017)}]{falke2017bringing}
Tobias Falke and Iryna Gurevych. 2017.
\newblock Bringing structure into summaries: Crowdsourcing a benchmark corpus
  of concept maps.
\newblock \emph{arXiv preprint arXiv:1704.04452}.

\bibitem[{Feigenblat et~al.(2021)Feigenblat, Gunasekara, Sznajder, Joshi,
  Konopnicki, and Aharonov}]{feigenblat-etal-2021-tweetsumm-dialog}
Guy Feigenblat, Chulaka Gunasekara, Benjamin Sznajder, Sachindra Joshi, David
  Konopnicki, and Ranit Aharonov. 2021.
\newblock \href {https://doi.org/10.18653/v1/2021.findings-emnlp.24}
  {{TWEETSUMM} - a dialog summarization dataset for customer service}.
\newblock In \emph{Findings of the Association for Computational Linguistics:
  EMNLP 2021}, pages 245--260, Punta Cana, Dominican Republic. Association for
  Computational Linguistics.

\bibitem[{Gehrmann et~al.(2018)Gehrmann, Deng, and
  Rush}]{gehrmann-etal-2018-bottom}
Sebastian Gehrmann, Yuntian Deng, and Alexander Rush. 2018.
\newblock \href {https://doi.org/10.18653/v1/D18-1443} {Bottom-up abstractive
  summarization}.
\newblock In \emph{Proceedings of the 2018 Conference on Empirical Methods in
  Natural Language Processing}, pages 4098--4109, Brussels, Belgium.
  Association for Computational Linguistics.

\bibitem[{Haviland and Clark(1974)}]{haviland1974s}
Susan~E Haviland and Herbert~H Clark. 1974.
\newblock What's new? acquiring new information as a process in comprehension.
\newblock \emph{Journal of verbal learning and verbal behavior},
  13(5):512--521.

\bibitem[{Hermann et~al.(2015)Hermann, Kocisky, Grefenstette, Espeholt, Kay,
  Suleyman, and Blunsom}]{hermann2015teaching}
Karl~Moritz Hermann, Tomas Kocisky, Edward Grefenstette, Lasse Espeholt, Will
  Kay, Mustafa Suleyman, and Phil Blunsom. 2015.
\newblock Teaching machines to read and comprehend.
\newblock In \emph{Advances in neural information processing systems}, pages
  1693--1701.

\bibitem[{Joshi et~al.(2020)Joshi, Chen, Liu, Weld, Zettlemoyer, and
  Levy}]{joshi-etal-2020-spanbert}
Mandar Joshi, Danqi Chen, Yinhan Liu, Daniel~S. Weld, Luke Zettlemoyer, and
  Omer Levy. 2020.
\newblock \href {https://doi.org/10.1162/tacl_a_00300} {{S}pan{BERT}: Improving
  pre-training by representing and predicting spans}.
\newblock \emph{Transactions of the Association for Computational Linguistics},
  8:64--77.

\bibitem[{Ju et~al.(2021)Ju, Liu, Koh, Jin, Du, and
  Pan}]{ju-etal-2021-leveraging-information}
Jiaxin Ju, Ming Liu, Huan~Yee Koh, Yuan Jin, Lan Du, and Shirui Pan. 2021.
\newblock \href {https://doi.org/10.18653/v1/2021.findings-emnlp.345}
  {Leveraging information bottleneck for scientific document summarization}.
\newblock In \emph{Findings of the Association for Computational Linguistics:
  EMNLP 2021}, pages 4091--4098, Punta Cana, Dominican Republic. Association
  for Computational Linguistics.

\bibitem[{Kintsch and Van~Dijk(1978)}]{kintsch1978toward}
Walter Kintsch and Teun~A Van~Dijk. 1978.
\newblock Toward a model of text comprehension and production.
\newblock \emph{Psychological review}, 85(5):363.

\bibitem[{Kitaev and Klein(2018)}]{kitaev2018constituency}
Nikita Kitaev and Dan Klein. 2018.
\newblock Constituency parsing with a self-attentive encoder.
\newblock \emph{arXiv preprint arXiv:1805.01052}.

\bibitem[{Koupaee and Wang(2018)}]{koupaee2018wikihow}
Mahnaz Koupaee and William~Yang Wang. 2018.
\newblock Wikihow: A large scale text summarization dataset.
\newblock \emph{arXiv preprint arXiv:1810.09305}.

\bibitem[{Lewis et~al.(2020)Lewis, Liu, Goyal, Ghazvininejad, Mohamed, Levy,
  Stoyanov, and Zettlemoyer}]{lewis-etal-2020-bart}
Mike Lewis, Yinhan Liu, Naman Goyal, Marjan Ghazvininejad, Abdelrahman Mohamed,
  Omer Levy, Veselin Stoyanov, and Luke Zettlemoyer. 2020.
\newblock \href {https://doi.org/10.18653/v1/2020.acl-main.703} {{BART}:
  Denoising sequence-to-sequence pre-training for natural language generation,
  translation, and comprehension}.
\newblock In \emph{Proceedings of the 58th Annual Meeting of the Association
  for Computational Linguistics}, pages 7871--7880, Online. Association for
  Computational Linguistics.

\bibitem[{Li et~al.(2021)Li, Ma, Yu, Wu, Gao, Ji, and
  McKeown}]{li-etal-2021-timeline}
Manling Li, Tengfei Ma, Mo~Yu, Lingfei Wu, Tian Gao, Heng Ji, and Kathleen
  McKeown. 2021.
\newblock \href {https://doi.org/10.18653/v1/2021.emnlp-main.519} {Timeline
  summarization based on event graph compression via time-aware optimal
  transport}.
\newblock In \emph{Proceedings of the 2021 Conference on Empirical Methods in
  Natural Language Processing}, pages 6443--6456, Online and Punta Cana,
  Dominican Republic. Association for Computational Linguistics.

\bibitem[{Lin(2004)}]{lin2004rouge}
Chin-Yew Lin. 2004.
\newblock Rouge: A package for automatic evaluation of summaries.
\newblock In \emph{Text summarization branches out}, pages 74--81.

\bibitem[{Liu et~al.(2021)Liu, Zou, Zhang, Chen, Ding, Yuan, and
  Wang}]{liu-etal-2021-topic-aware}
Junpeng Liu, Yanyan Zou, Hainan Zhang, Hongshen Chen, Zhuoye Ding, Caixia Yuan,
  and Xiaojie Wang. 2021.
\newblock \href {https://doi.org/10.18653/v1/2021.findings-emnlp.106}
  {Topic-aware contrastive learning for abstractive dialogue summarization}.
\newblock In \emph{Findings of the Association for Computational Linguistics:
  EMNLP 2021}, pages 1229--1243, Punta Cana, Dominican Republic. Association
  for Computational Linguistics.

\bibitem[{Liu and Lapata(2019)}]{liu2019text}
Yang Liu and Mirella Lapata. 2019.
\newblock Text summarization with pretrained encoders.
\newblock \emph{arXiv preprint arXiv:1908.08345}.

\bibitem[{Luan et~al.(2019)Luan, Wadden, He, Shah, Ostendorf, and
  Hajishirzi}]{luan2019general}
Yi~Luan, Dave Wadden, Luheng He, Amy Shah, Mari Ostendorf, and Hannaneh
  Hajishirzi. 2019.
\newblock A general framework for information extraction using dynamic span
  graphs.
\newblock \emph{arXiv preprint arXiv:1904.03296}.

\bibitem[{Mani(2001)}]{mani2001automatic}
Inderjeet Mani. 2001.
\newblock \emph{Automatic summarization}, volume~3.
\newblock John Benjamins Publishing.

\bibitem[{Maynez et~al.(2020)Maynez, Narayan, Bohnet, and
  McDonald}]{maynez-etal-2020-faithfulness}
Joshua Maynez, Shashi Narayan, Bernd Bohnet, and Ryan McDonald. 2020.
\newblock \href {https://doi.org/10.18653/v1/2020.acl-main.173} {On
  faithfulness and factuality in abstractive summarization}.
\newblock In \emph{Proceedings of the 58th Annual Meeting of the Association
  for Computational Linguistics}, pages 1906--1919, Online. Association for
  Computational Linguistics.

\bibitem[{Meng et~al.(2021)Meng, Thaker, Zhang, Dong, Yuan, Wang, and
  He}]{meng-etal-2021-bringing}
Rui Meng, Khushboo Thaker, Lei Zhang, Yue Dong, Xingdi Yuan, Tong Wang, and
  Daqing He. 2021.
\newblock \href {https://doi.org/10.18653/v1/2021.acl-short.137} {Bringing
  structure into summaries: a faceted summarization dataset for long scientific
  documents}.
\newblock In \emph{Proceedings of the 59th Annual Meeting of the Association
  for Computational Linguistics and the 11th International Joint Conference on
  Natural Language Processing (Volume 2: Short Papers)}, pages 1080--1089,
  Online. Association for Computational Linguistics.

\bibitem[{Narayan et~al.(2018{\natexlab{a}})Narayan, Cohen, and
  Lapata}]{narayan2018don}
Shashi Narayan, Shay~B Cohen, and Mirella Lapata. 2018{\natexlab{a}}.
\newblock Don't give me the details, just the summary! topic-aware
  convolutional neural networks for extreme summarization.
\newblock \emph{arXiv preprint arXiv:1808.08745}.

\bibitem[{Narayan et~al.(2018{\natexlab{b}})Narayan, Cohen, and
  Lapata}]{narayan-etal-2018-ranking}
Shashi Narayan, Shay~B. Cohen, and Mirella Lapata. 2018{\natexlab{b}}.
\newblock \href {https://doi.org/10.18653/v1/N18-1158} {Ranking sentences for
  extractive summarization with reinforcement learning}.
\newblock In \emph{Proceedings of the 2018 Conference of the North {A}merican
  Chapter of the Association for Computational Linguistics: Human Language
  Technologies, Volume 1 (Long Papers)}, pages 1747--1759, New Orleans,
  Louisiana. Association for Computational Linguistics.

\bibitem[{Parton et~al.(2009)Parton, McKeown, Coyne, Diab, Grishman,
  Hakkani-T{\"u}r, Harper, Ji, Ma, Meyers et~al.}]{parton2009comparing}
Kristen Parton, Kathleen McKeown, Robert~Eric Coyne, Mona~T Diab, Ralph
  Grishman, Dilek Hakkani-T{\"u}r, Mary Harper, Heng Ji, Wei~Yun Ma, Adam
  Meyers, et~al. 2009.
\newblock Who, what, when, where, why? comparing multiple approaches to the
  cross-lingual 5w task.

\bibitem[{Paulus et~al.(2017)Paulus, Xiong, and Socher}]{paulus2017deep}
Romain Paulus, Caiming Xiong, and Richard Socher. 2017.
\newblock A deep reinforced model for abstractive summarization.
\newblock \emph{arXiv preprint arXiv:1705.04304}.

\bibitem[{Qian et~al.(2018)Qian, Santus, Jin, Guo, and
  Barzilay}]{qian2018graphie}
Yujie Qian, Enrico Santus, Zhijing Jin, Jiang Guo, and Regina Barzilay. 2018.
\newblock Graphie: A graph-based framework for information extraction.
\newblock \emph{arXiv preprint arXiv:1810.13083}.

\bibitem[{Raffel et~al.(2019)Raffel, Shazeer, Roberts, Lee, Narang, Matena,
  Zhou, Li, and Liu}]{raffel2019exploring}
Colin Raffel, Noam Shazeer, Adam Roberts, Katherine Lee, Sharan Narang, Michael
  Matena, Yanqi Zhou, Wei Li, and Peter~J Liu. 2019.
\newblock Exploring the limits of transfer learning with a unified text-to-text
  transformer.
\newblock \emph{arXiv preprint arXiv:1910.10683}.

\bibitem[{Ratner et~al.(2017)Ratner, Bach, Ehrenberg, Fries, Wu, and
  R{\'e}}]{ratner2017snorkel}
Alexander Ratner, Stephen~H Bach, Henry Ehrenberg, Jason Fries, Sen Wu, and
  Christopher R{\'e}. 2017.
\newblock Snorkel: Rapid training data creation with weak supervision.
\newblock In \emph{Proceedings of the VLDB Endowment. International Conference
  on Very Large Data Bases}, volume~11, page 269. NIH Public Access.

\bibitem[{Ribeiro et~al.(2022)Ribeiro, Liu, Gurevych, Dreyer, and
  Bansal}]{ribeiro2022fact}
Leonardo F.~R. Ribeiro, Mengwen Liu, Iryna Gurevych, Markus Dreyer, and Mohit
  Bansal. 2022.
\newblock Factgraph: Evaluating factuality in summarization with semantic graph
  representations.
\newblock \emph{arXiv preprint arXiv:2204.06508}.

\bibitem[{See et~al.(2017)See, Liu, and Manning}]{see2017get}
Abigail See, Peter~J Liu, and Christopher~D Manning. 2017.
\newblock Get to the point: Summarization with pointer-generator networks.
\newblock \emph{arXiv preprint arXiv:1704.04368}.

\bibitem[{Sp{\"a}rck~Jones(1998)}]{jones1999automatic}
Karen Sp{\"a}rck~Jones. 1998.
\newblock Automatic summarizing: factors and directions.
\newblock In \emph{Advances in automatic text summarization}, 1, pages 1--12.
  MIT press Cambridge, Mass, USA.

\bibitem[{ter Hoeve et~al.(2020)ter Hoeve, Kiseleva, and
  de~Rijke}]{ter2020makes}
Maartje ter Hoeve, Julia Kiseleva, and Maarten de~Rijke. 2020.
\newblock What makes a good summary? {Reconsidering} the focus of automatic
  summarization.
\newblock \emph{arXiv preprint arXiv:2012.07619}.

\bibitem[{Vaswani et~al.(2017)Vaswani, Shazeer, Parmar, Uszkoreit, Jones,
  Gomez, Kaiser, and Polosukhin}]{vaswani2017attention}
Ashish Vaswani, Noam Shazeer, Niki Parmar, Jakob Uszkoreit, Llion Jones,
  Aidan~N Gomez, {\L}ukasz Kaiser, and Illia Polosukhin. 2017.
\newblock Attention is all you need.
\newblock \emph{Advances in neural information processing systems}, 30.

\bibitem[{Wadden et~al.(2019)Wadden, Wennberg, Luan, and
  Hajishirzi}]{wadden2019entity}
David Wadden, Ulme Wennberg, Yi~Luan, and Hannaneh Hajishirzi. 2019.
\newblock Entity, relation, and event extraction with contextualized span
  representations.
\newblock \emph{arXiv preprint arXiv:1909.03546}.

\bibitem[{Wang et~al.(2020)Wang, Cho, and Lewis}]{wang2020asking}
Alex Wang, Kyunghyun Cho, and Mike Lewis. 2020.
\newblock Asking and answering questions to evaluate the factual consistency of
  summaries.
\newblock \emph{arXiv preprint arXiv:2004.04228}.

\bibitem[{Wolf et~al.(2020)Wolf, Debut, Sanh, Chaumond, Delangue, Moi, Cistac,
  Rault, Louf, Funtowicz, Davison, Shleifer, von Platen, Ma, Jernite, Plu, Xu,
  Le~Scao, Gugger, Drame, Lhoest, and Rush}]{wolf-etal-2020-transformers}
Thomas Wolf, Lysandre Debut, Victor Sanh, Julien Chaumond, Clement Delangue,
  Anthony Moi, Pierric Cistac, Tim Rault, Remi Louf, Morgan Funtowicz, Joe
  Davison, Sam Shleifer, Patrick von Platen, Clara Ma, Yacine Jernite, Julien
  Plu, Canwen Xu, Teven Le~Scao, Sylvain Gugger, Mariama Drame, Quentin Lhoest,
  and Alexander Rush. 2020.
\newblock \href {https://doi.org/10.18653/v1/2020.emnlp-demos.6} {Transformers:
  State-of-the-art natural language processing}.
\newblock In \emph{Proceedings of the 2020 Conference on Empirical Methods in
  Natural Language Processing: System Demonstrations}, pages 38--45, Online.
  Association for Computational Linguistics.

\bibitem[{Wu et~al.(2020)Wu, Koncel-Kedziorski, Ostendorf, and
  Hajishirzi}]{wu2020extracting}
Zeqiu Wu, Rik Koncel-Kedziorski, Mari Ostendorf, and Hannaneh Hajishirzi. 2020.
\newblock Extracting summary knowledge graphs from long documents.
\newblock \emph{arXiv preprint arXiv:2009.09162}.

\bibitem[{Xue et~al.(2021)Xue, Constant, Roberts, Kale, Al-Rfou, Siddhant,
  Barua, and Raffel}]{xue-etal-2021-mt5}
Linting Xue, Noah Constant, Adam Roberts, Mihir Kale, Rami Al-Rfou, Aditya
  Siddhant, Aditya Barua, and Colin Raffel. 2021.
\newblock \href {https://doi.org/10.18653/v1/2021.naacl-main.41} {m{T}5: A
  massively multilingual pre-trained text-to-text transformer}.
\newblock In \emph{Proceedings of the 2021 Conference of the North American
  Chapter of the Association for Computational Linguistics: Human Language
  Technologies}, pages 483--498, Online. Association for Computational
  Linguistics.

\bibitem[{Yu et~al.(2021)Yu, Jatowt, Doucet, Sugiyama, and
  Yoshikawa}]{yu-etal-2021-multi}
Yi~Yu, Adam Jatowt, Antoine Doucet, Kazunari Sugiyama, and Masatoshi Yoshikawa.
  2021.
\newblock \href {https://doi.org/10.18653/v1/2021.acl-long.32}
  {Multi-{T}ime{L}ine summarization ({MTLS}): Improving timeline summarization
  by generating multiple summaries}.
\newblock In \emph{Proceedings of the 59th Annual Meeting of the Association
  for Computational Linguistics and the 11th International Joint Conference on
  Natural Language Processing (Volume 1: Long Papers)}, pages 377--387, Online.
  Association for Computational Linguistics.

\end{thebibliography}
\bibliographystyle{acl_natbib}

\appendix
\onecolumn
\section*{Appendix}

%!TEX root = ../main.tex

\section{Human labeling}
\label{sec:appendix_human_labeling}

\begin{figure*}[htp!]
    \centering
    \begin{subfigure}{\textwidth}
        \centering
        \includegraphics[width=\textwidth]{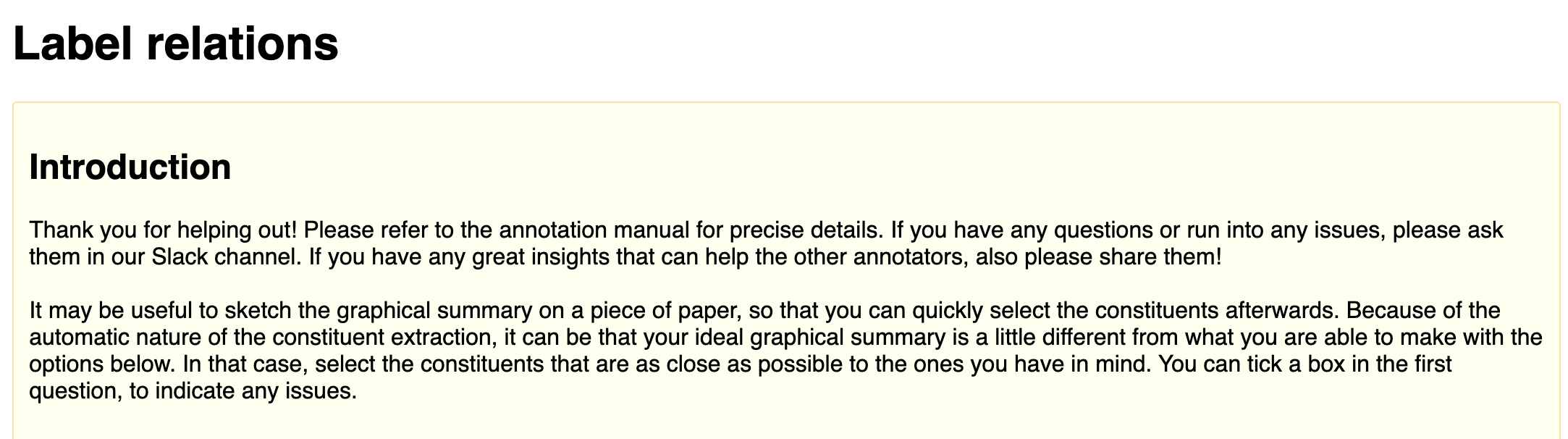}
    \end{subfigure}
    \begin{subfigure}{\textwidth}
        \centering
        \includegraphics[width=\textwidth]{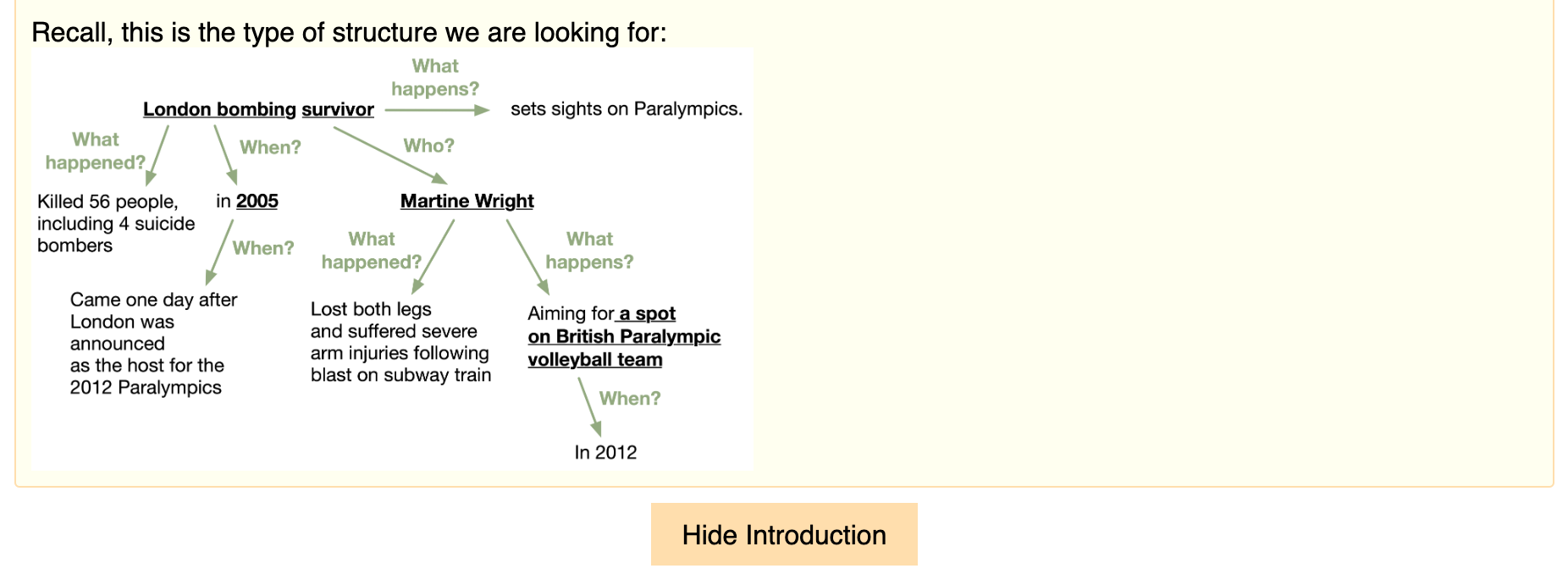}
    \end{subfigure}
    \begin{subfigure}{\textwidth}
        \centering
        \includegraphics[width=\textwidth]{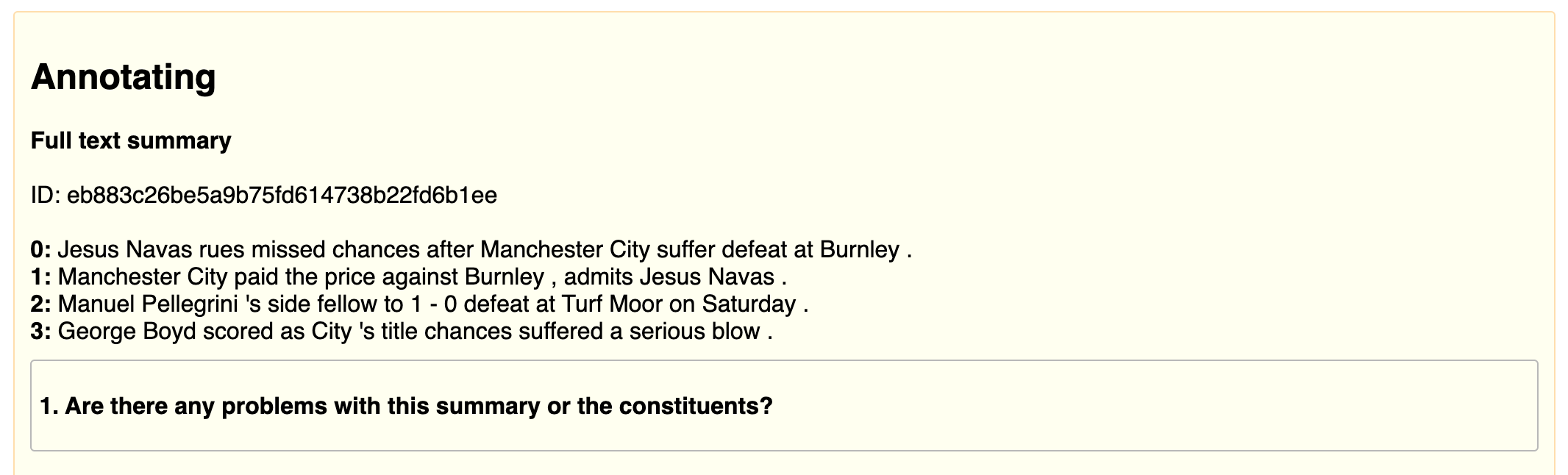}
    \end{subfigure}
    \begin{subfigure}{\textwidth}
        \centering
        \includegraphics[width=\textwidth]{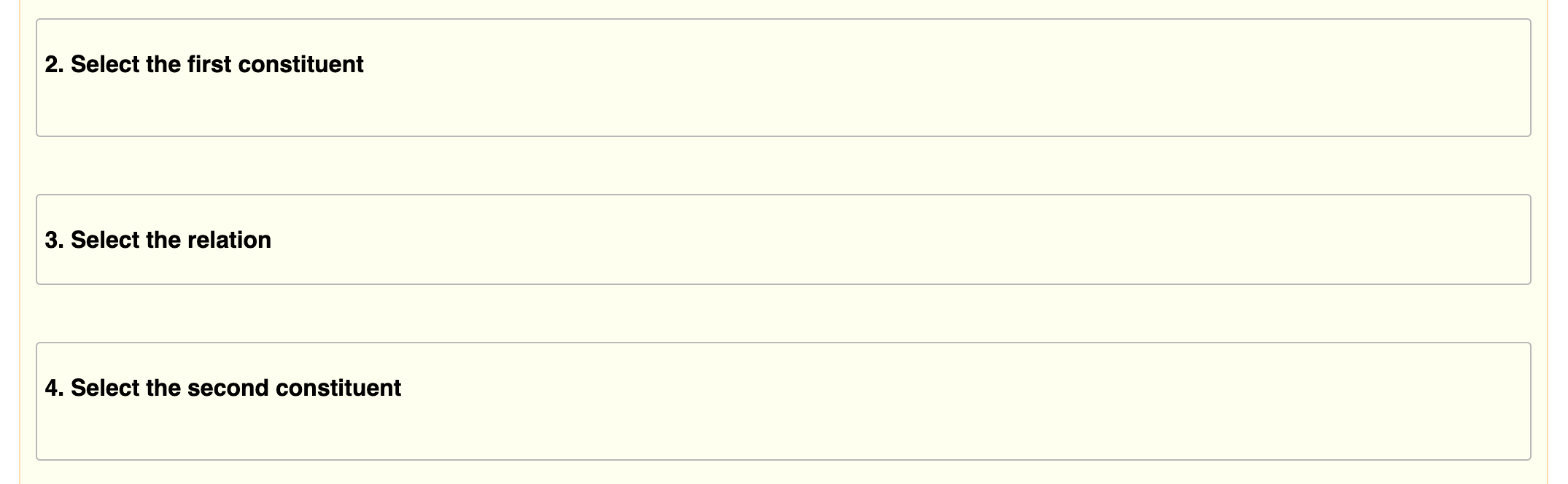}
    \end{subfigure}
    \begin{subfigure}{\textwidth}
        \centering
        \includegraphics[width=\textwidth]{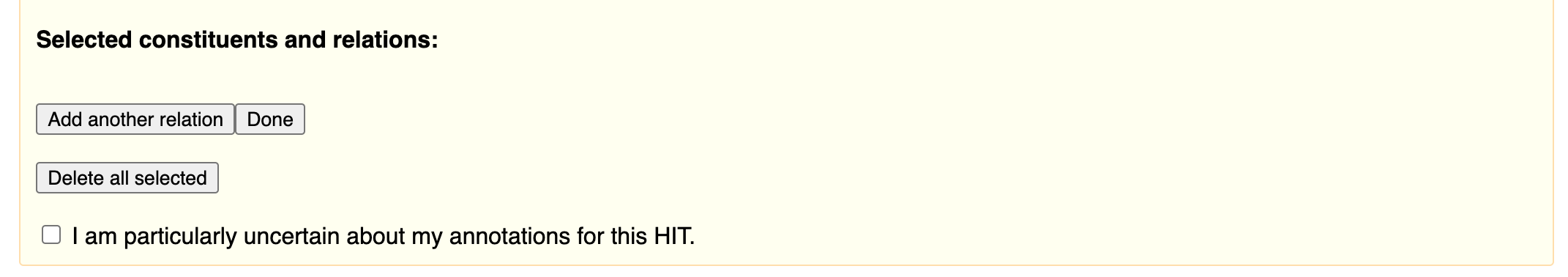}
    \end{subfigure}
  \centering
  \caption{Overview of human labeling task. Figure consists of several screen shots of the task.}
\label{fig:human_labeling_overview}
\vspace{20em}
\end{figure*}

\begin{figure*}[ht!]
    \centering
    \begin{subfigure}{\textwidth}
        \centering
        \includegraphics[width=\textwidth]{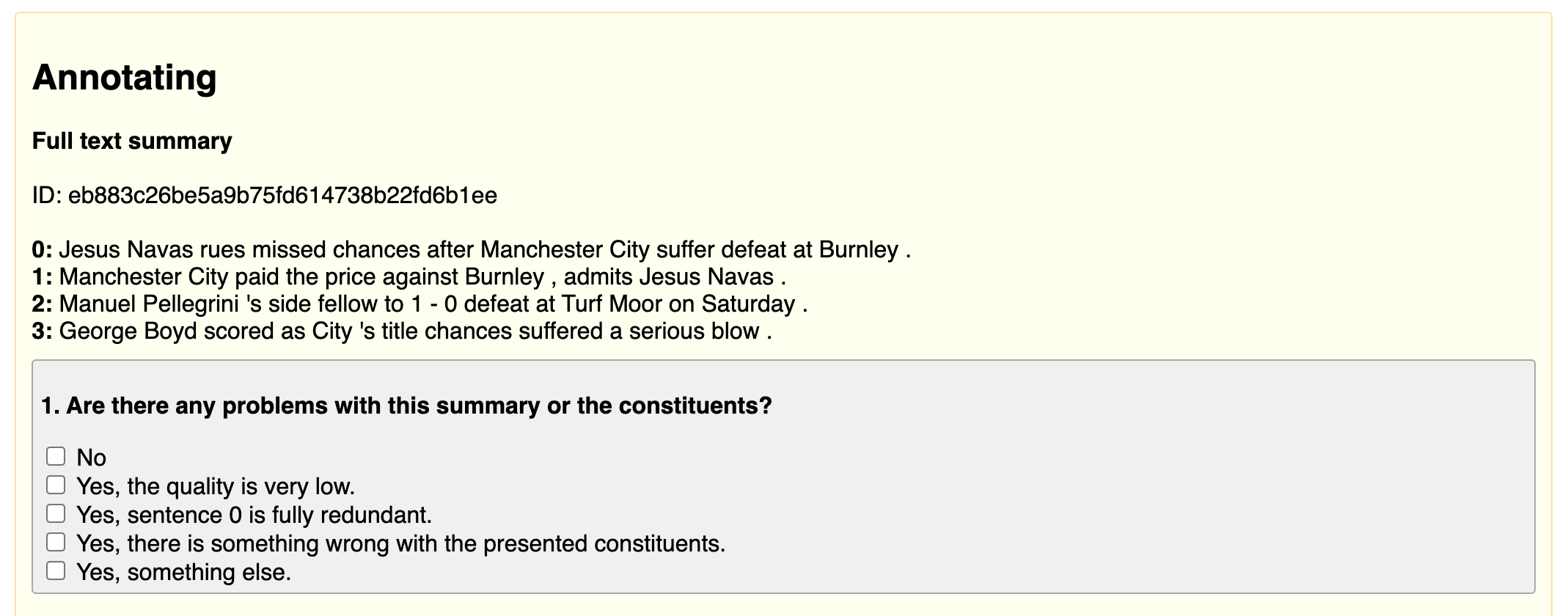}
        \caption{Expansion of question 1.}
    \end{subfigure}
    \begin{subfigure}{\textwidth}
        \centering
        \includegraphics[width=\textwidth]{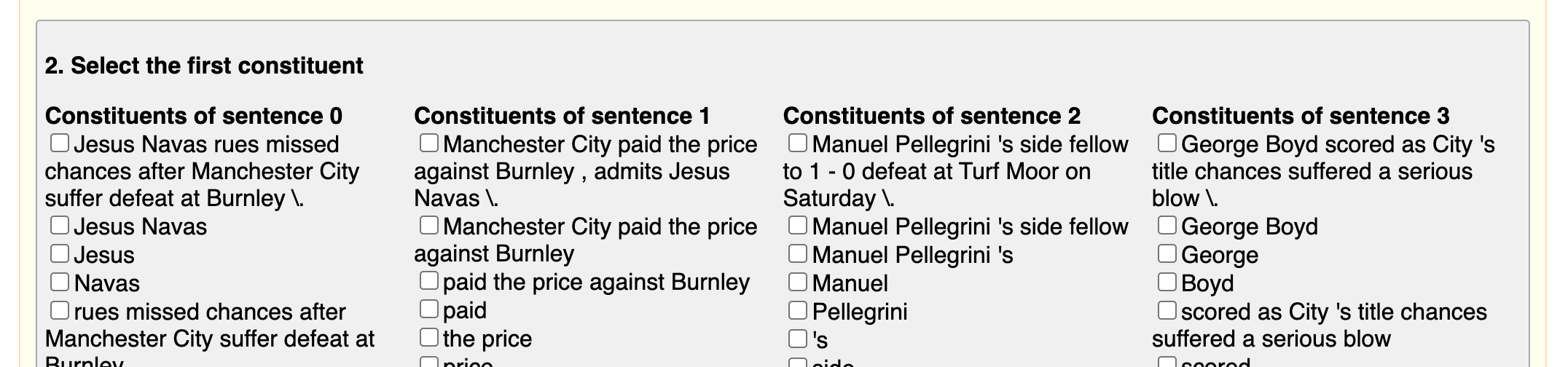}
        \caption{Partial expansion of selecting constituents.}
    \end{subfigure}
      \begin{subfigure}{\textwidth}
        \centering
        \includegraphics[width=\textwidth]{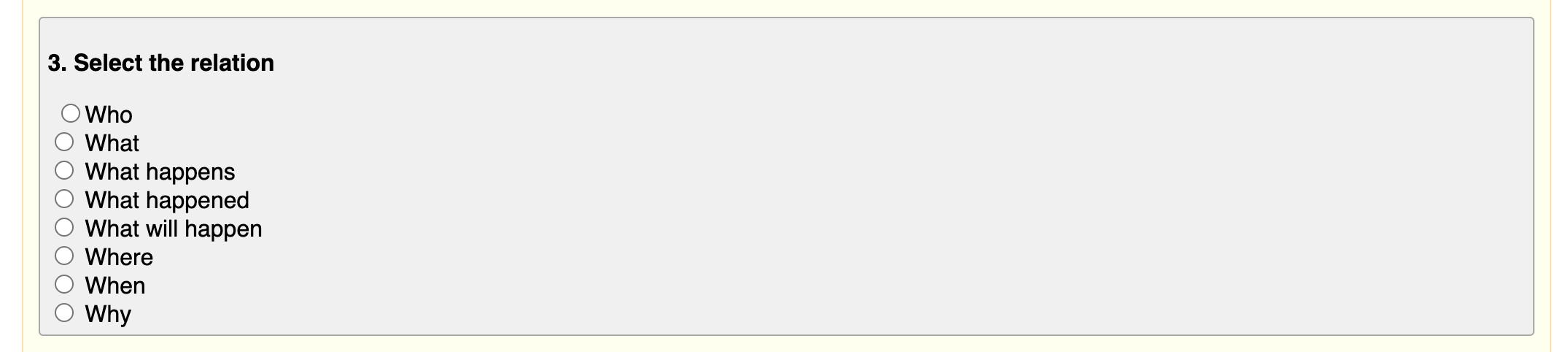}
        \caption{Expansion of selecting relations.}
    \end{subfigure}
  \centering
  \caption{Human labeling task in more detail.}
  \vspace{10pt}
  \label{fig:human_labeling_details}
  
\end{figure*}

Figure~\ref{fig:human_labeling_overview} and~\ref{fig:human_labeling_details} show screen shots of the human labeling task.

%!TEX root = ../main.tex

\section{Q1 -- Human evaluation}
\label{sec:appendix_human_eval_q1}

\subsection{Examples of task setup}
\label{sec:appendix_human_eval_q1_task_setup}

\begin{figure*}
    \centering
    \begin{subfigure}{\textwidth}
        \centering
        \includegraphics[width=\textwidth]{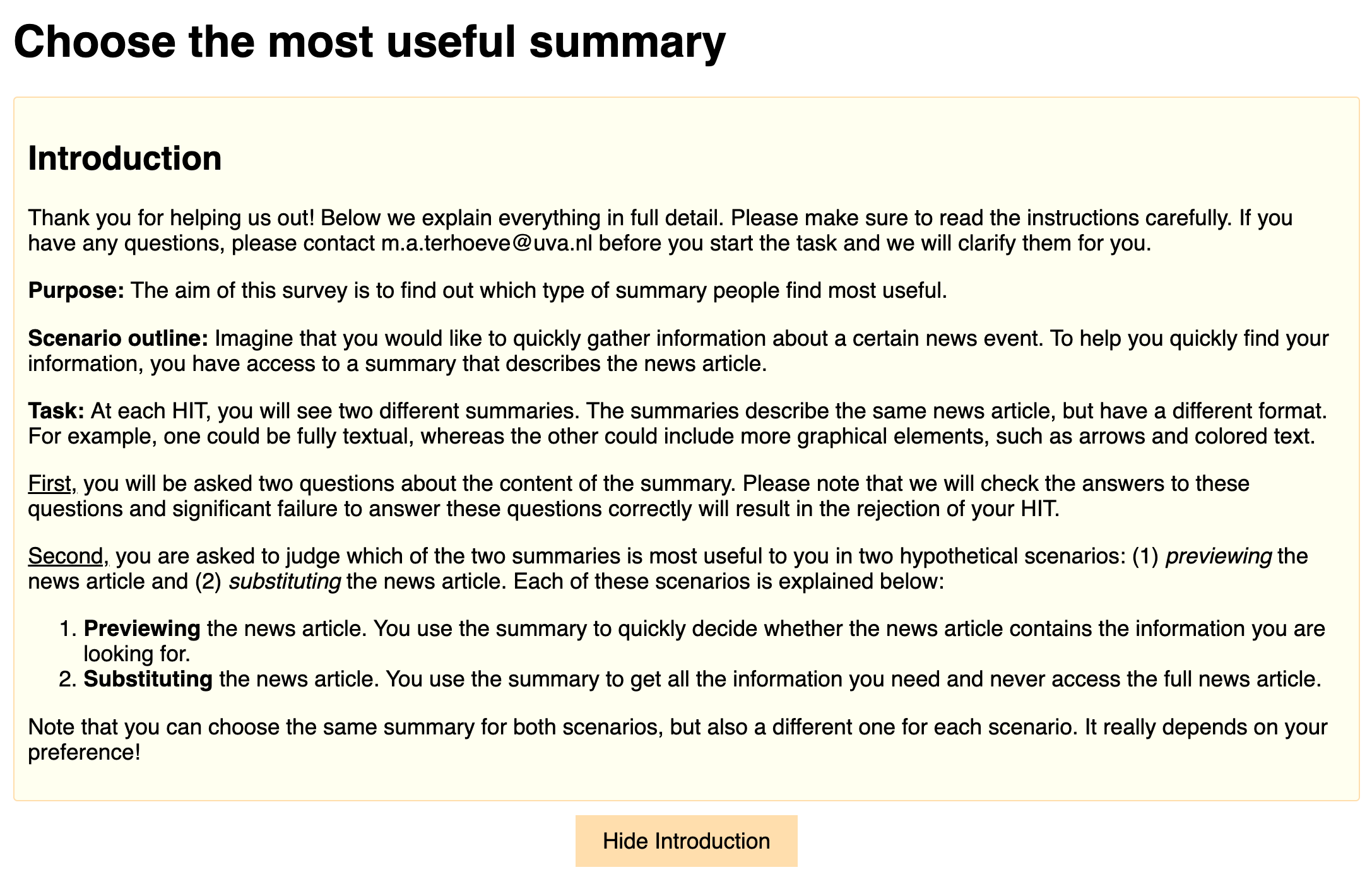}
    \end{subfigure}
    \begin{subfigure}{\textwidth}
        \centering
        \includegraphics[width=\textwidth]{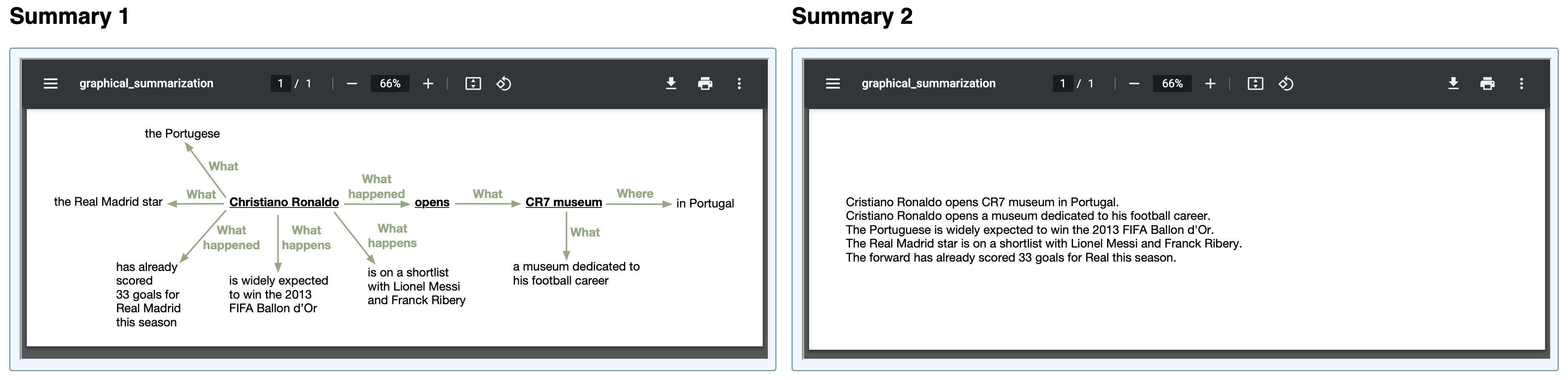}
    \end{subfigure}
        \begin{subfigure}{\textwidth}
        \centering
        \includegraphics[width=\textwidth]{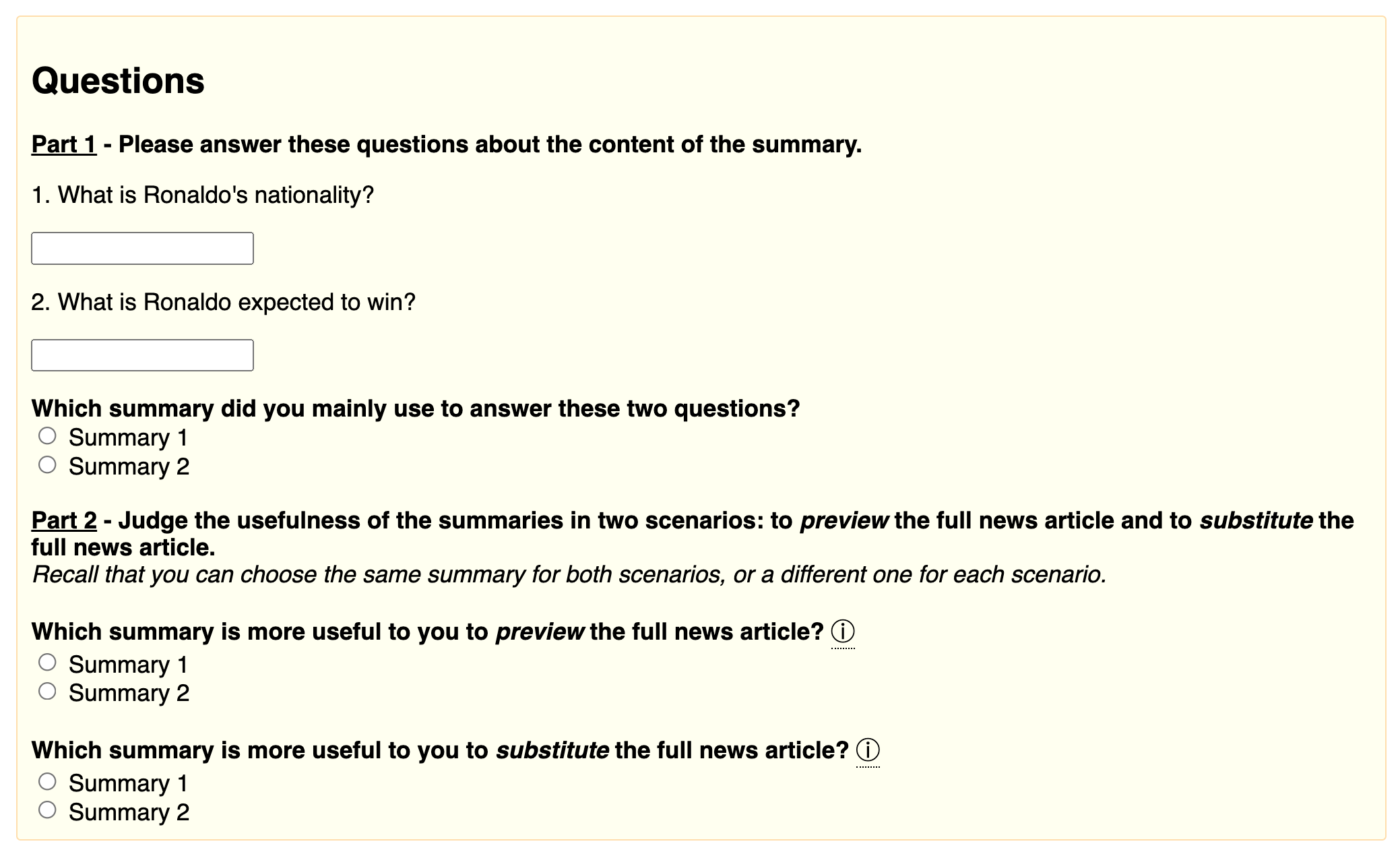}
    \end{subfigure}
    \caption{Example of human evaluation task. These are three screen shots -- Summary 1 and 2 have a larger font in the actual task interface so they are more easily readable for workers.}
    \label{fig:human_eval_q1_overview}

  \centering
\end{figure*}

Figure \ref{fig:human_eval_q1_overview} shows screen shots of the human evaluation task setup for question 1, where we investigate whether a critical mass of people is interested in summaries with graphical elements.

\twocolumn
\subsection{Additional results human evaluation Q1}
\label{sec:appendix_human_eval_q1_additional_results}

In Table~\ref{tab:human_eval_q1_doc1} and~\ref{tab:human_eval_q1_doc2} we give the results of the human evaluation per document.

\begin{table}
\begin{tabular}{llrr}
\toprule
	& \multicolumn{1}{c}{\textbf{Pair (A/B)}} & \multicolumn{1}{c}{\textbf{Prefer A}} & \multicolumn{1}{c}{\textbf{Prefer B}} \\
	& \multicolumn{1}{c}{} & \multicolumn{1}{c}{\textbf{(\%)}} & \multicolumn{1}{c}{\textbf{(\%)}}\\
	\midrule
\parbox[t]{2mm}{\multirow{3}{*}{\rotatebox[origin=c]{90}{\emph{Used}}}}
        &  Graphical / Text  & $40.0$ & $60.0$  \\
        &  Graphical / Typeset  & $50.0$ & $50.0$  \\
	      &  Typeset / Text  & $85.0$ & $15.0$ \\ \hline
\parbox[t]{2mm}{\multirow{3}{*}{\rotatebox[origin=c]{90}{\emph{Prev}}}}
	&  Graphical / Text   & $35.0$  & $65.0$ \\
	&  Graphical / Typeset  & $55.0$ & $45.0$  \\
	&  Typeset / Text   & $90.0$ & $10.0$ \\ \hline
\parbox[t]{2mm}{\multirow{3}{*}{\rotatebox[origin=c]{90}{\emph{Sub}}}}
	&  Graphical / Text  & $20.0$ & $80.0$ \\
	&  Graphical / Typeset  &$30.0$  & $70.0$ \\
	&  Typeset / Text  & $75.0$ & $25.0$  \\
	\bottomrule
\end{tabular}
\caption{Results Human Evaluation. Pairwise comparisons. Results for Document $1$.}
\label{tab:human_eval_q1_doc1}
\end{table}

\begin{table}
\begin{tabular}{llrr}
\toprule
	& \multicolumn{1}{c}{\textbf{Pair (A/B)}} & \multicolumn{1}{c}{\textbf{Prefer A}} & \multicolumn{1}{c}{\textbf{Prefer B}} \\
	& \multicolumn{1}{c}{} & \multicolumn{1}{c}{\textbf{(\%)}} & \multicolumn{1}{c}{\textbf{(\%)}}\\
	\midrule
\parbox[t]{2mm}{\multirow{3}{*}{\rotatebox[origin=c]{90}{\emph{Used}}}}
        &  Graphical / Text  & $30.0$ & $70.0$  \\
        &  Graphical / Typeset  & $50.0$ & $50.0$  \\
	      &  Typeset / Text  & $60.0$ & $40.0$ \\ \hline
\parbox[t]{2mm}{\multirow{3}{*}{\rotatebox[origin=c]{90}{\emph{Prev}}}}
	&  Graphical / Text   & $35.0$  & $65.0$ \\
	&  Graphical / Typeset  & $40.0$ & $60.0$  \\
	&  Typeset / Text   & $60.0$ & $40.0$ \\ \hline
\parbox[t]{2mm}{\multirow{3}{*}{\rotatebox[origin=c]{90}{\emph{Sub}}}}
	&  Graphical / Text  & $40.0$ & $60.0$ \\
	&  Graphical / Typeset  & $35.0$  & $65.0$ \\
	&  Typeset / Text  & $55.0$ & $45.0$  \\
	\bottomrule
\end{tabular}
\caption{Results Human Evaluation. Pairwise comparisons. Results for Document $2$.}
\label{tab:human_eval_q1_doc2}
\end{table}

\onecolumn
\subsection{List of open questions}
\label{sec:open_questions_human_eval_q1}

\paragraph{Document 1}
\begin{enumerate}[leftmargin=*,nosep]
    \item How expensive was the replica of the ark?
    \item What is the name of the carpenter?
    \item Where is the replica of the ark?
    \item How long did Johan Huyberts spend on building the ark?
    \item What did the carpenter dream?
    \item What can visitors do in the ark?
\end{enumerate}

\paragraph{Document 2}
\begin{enumerate}[leftmargin=*,nosep]
    \item What is Ronaldo's nationality?
    \item What is Ronaldo expected to win?
    \item Where did Ronaldo open a museum?
    \item How many goals has Ronaldo scored for Real Madrid this season?
    \item What is the CR7 museum?
    \item Who are on the short list together with Ronaldo?
\end{enumerate}
%!TEX root = ../main.tex

\section{Q2 -- Human evaluation}
\label{sec:appendix_human_eval_q2}

\subsection{Examples of task setup}

In Figure~\ref{fig:human_eval_q2_overview} we show screen shots of the task setup for the human evaluation where we compare the results of different methods on our task.

\begin{figure*}[h]
    \centering
    \begin{subfigure}{\textwidth}
        \centering
        \includegraphics[width=\textwidth]{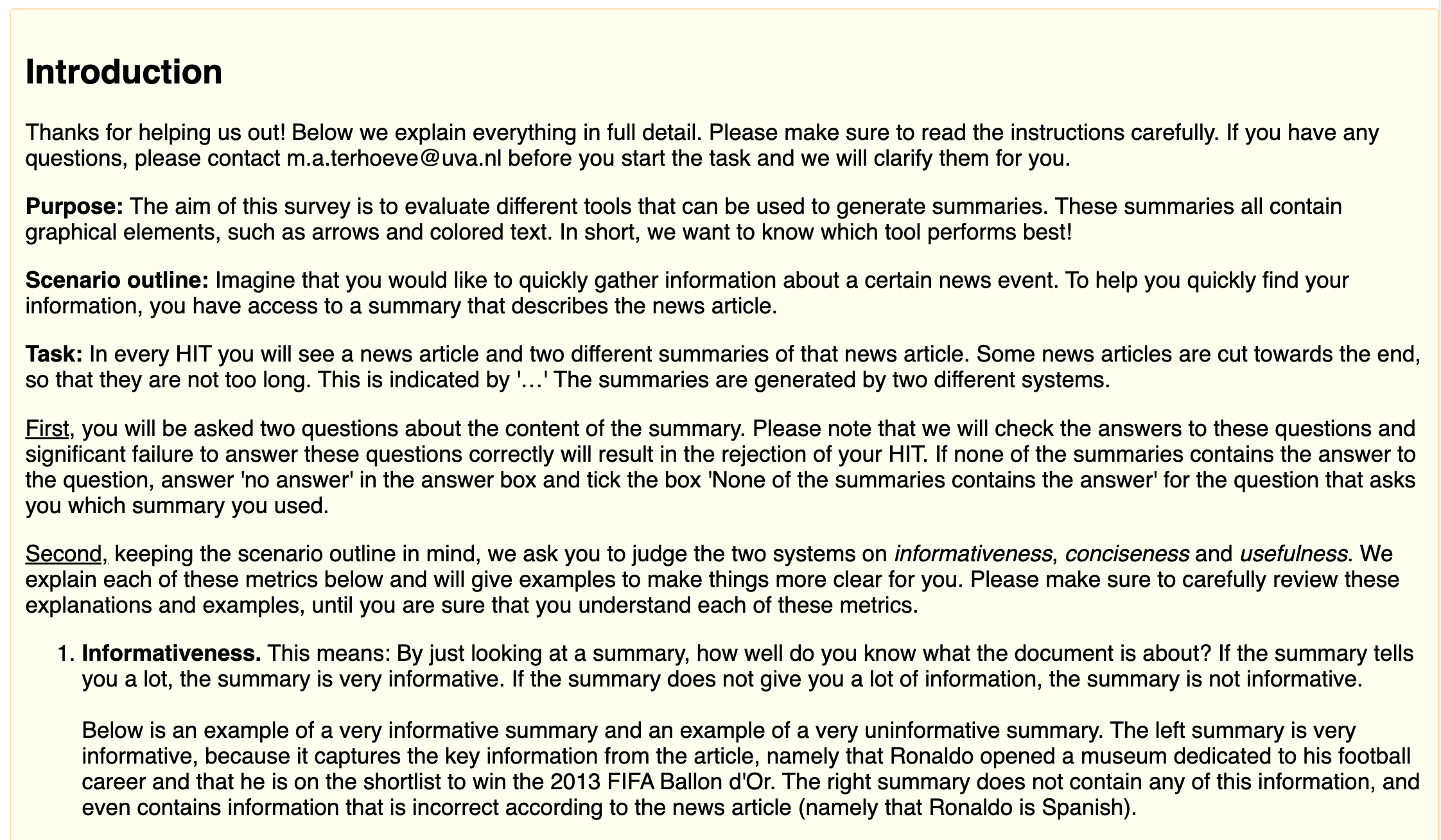}
    \end{subfigure}
    \begin{subfigure}{\textwidth}
        \centering
        \includegraphics[width=\textwidth]{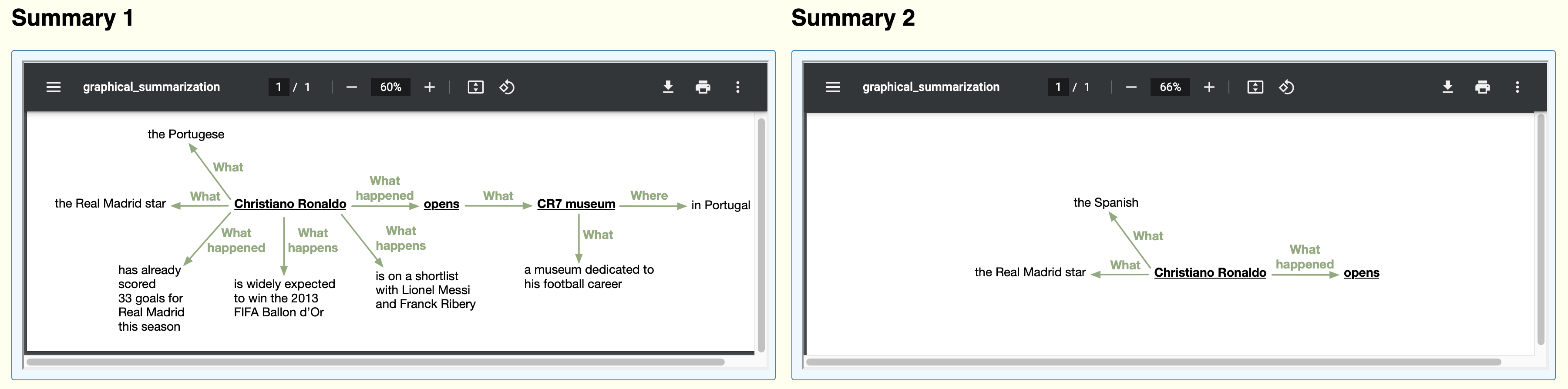}
    \end{subfigure}
    \begin{subfigure}{\textwidth}
        \centering
        \includegraphics[width=\textwidth]{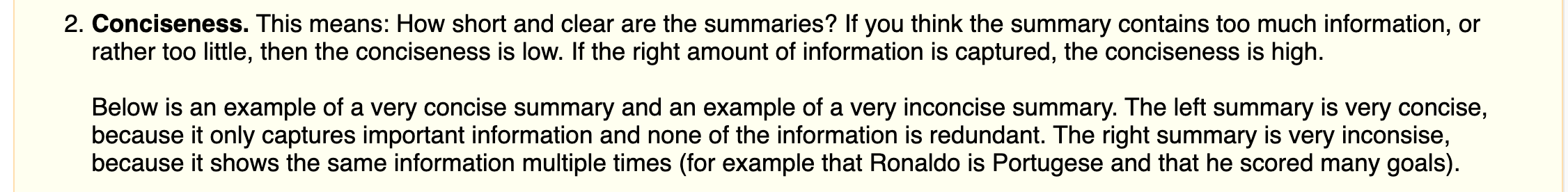}
    \end{subfigure}
    \begin{subfigure}{\textwidth}
        \centering
        \includegraphics[width=\textwidth]{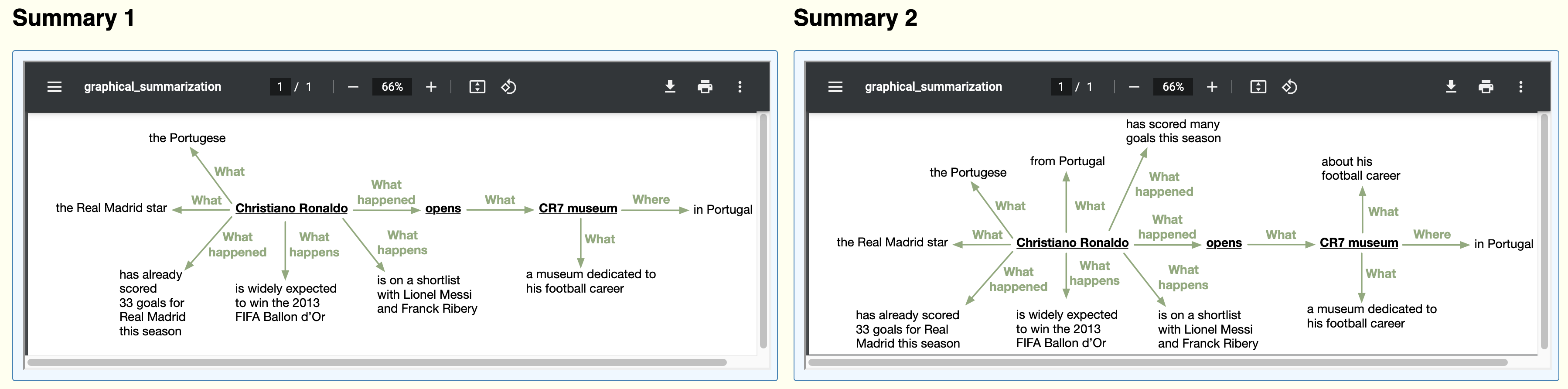}
    \end{subfigure}
    \begin{subfigure}{\textwidth}
        \centering
        \includegraphics[width=\textwidth]{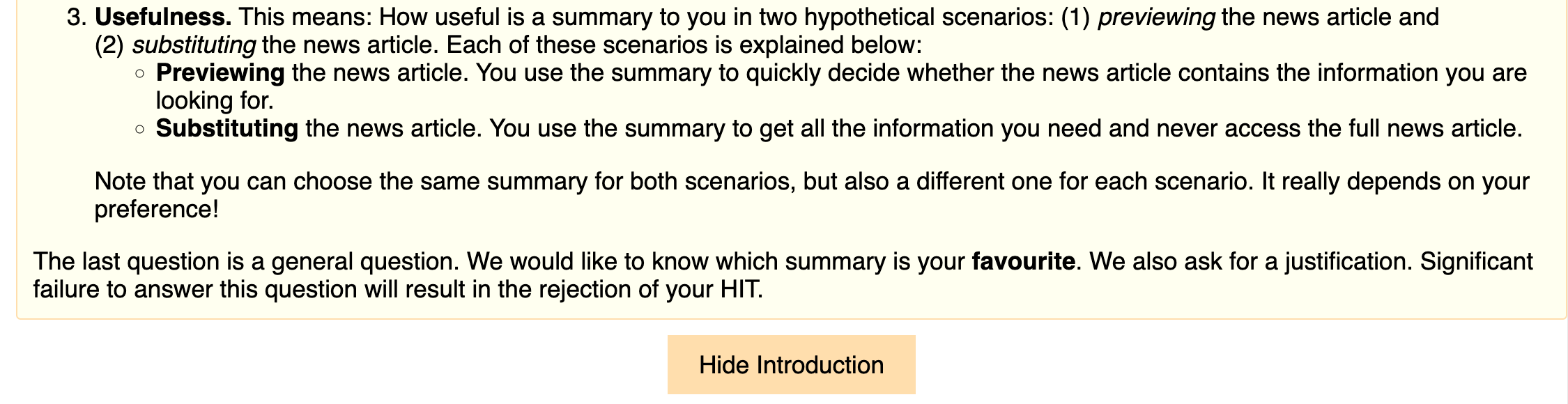}
    \end{subfigure}
\label{fig:human_eval_q2_overview}
  \centering
\end{figure*}

\begin{figure*}[ht!]\ContinuedFloat
    \centering
    \begin{subfigure}{\textwidth}
        \centering
        \includegraphics[width=\textwidth]{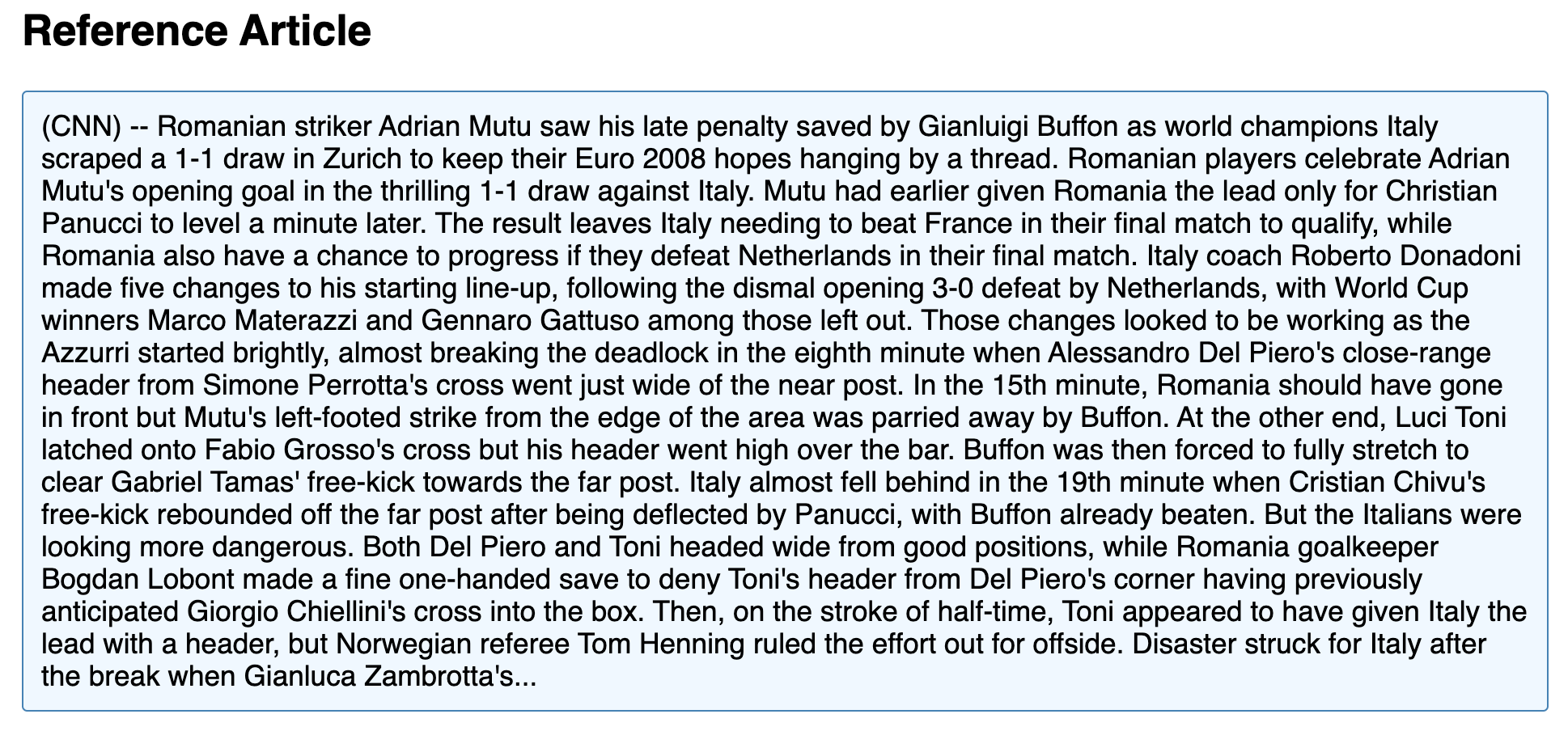}
    \end{subfigure}
    \begin{subfigure}{\textwidth}
        \centering
        \includegraphics[width=\textwidth]{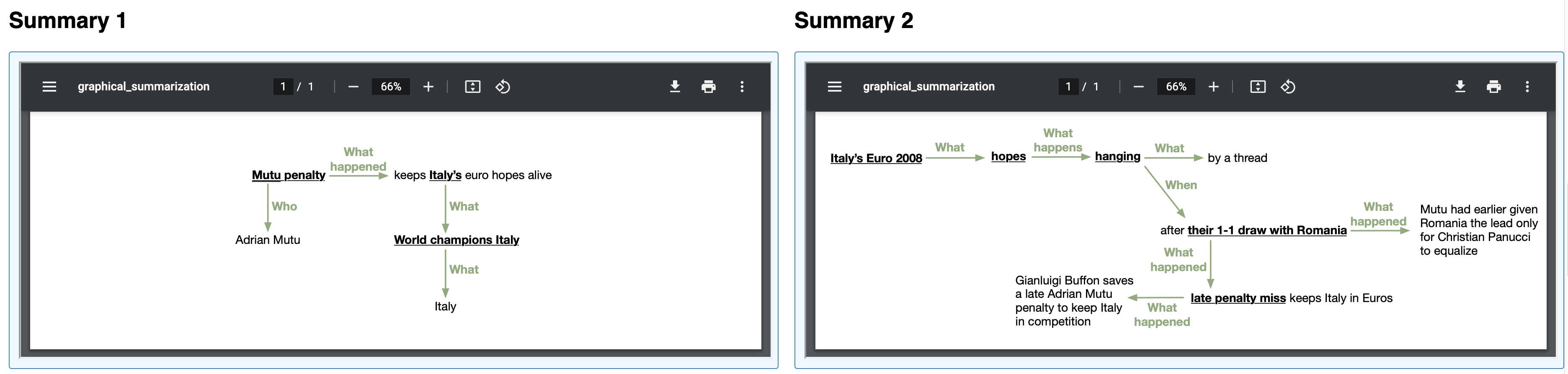}
    \end{subfigure}
    \begin{subfigure}{\textwidth}
        \centering
        \includegraphics[width=\textwidth]{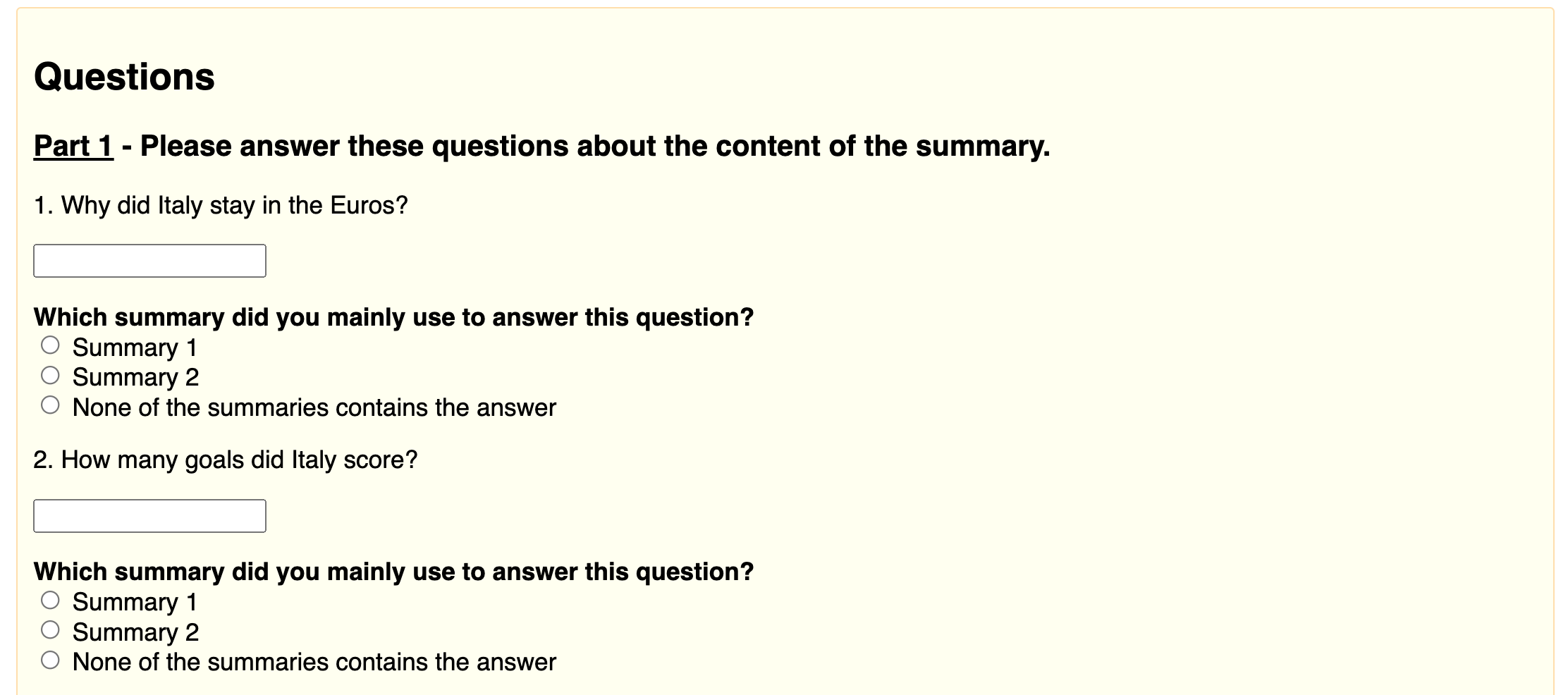}
    \end{subfigure}
\label{fig:human_eval_q2_overview}
  \centering
\end{figure*}

\begin{figure*}[ht!]\ContinuedFloat
    \centering
    \begin{subfigure}{\textwidth}
        \centering
        \includegraphics[width=\textwidth]{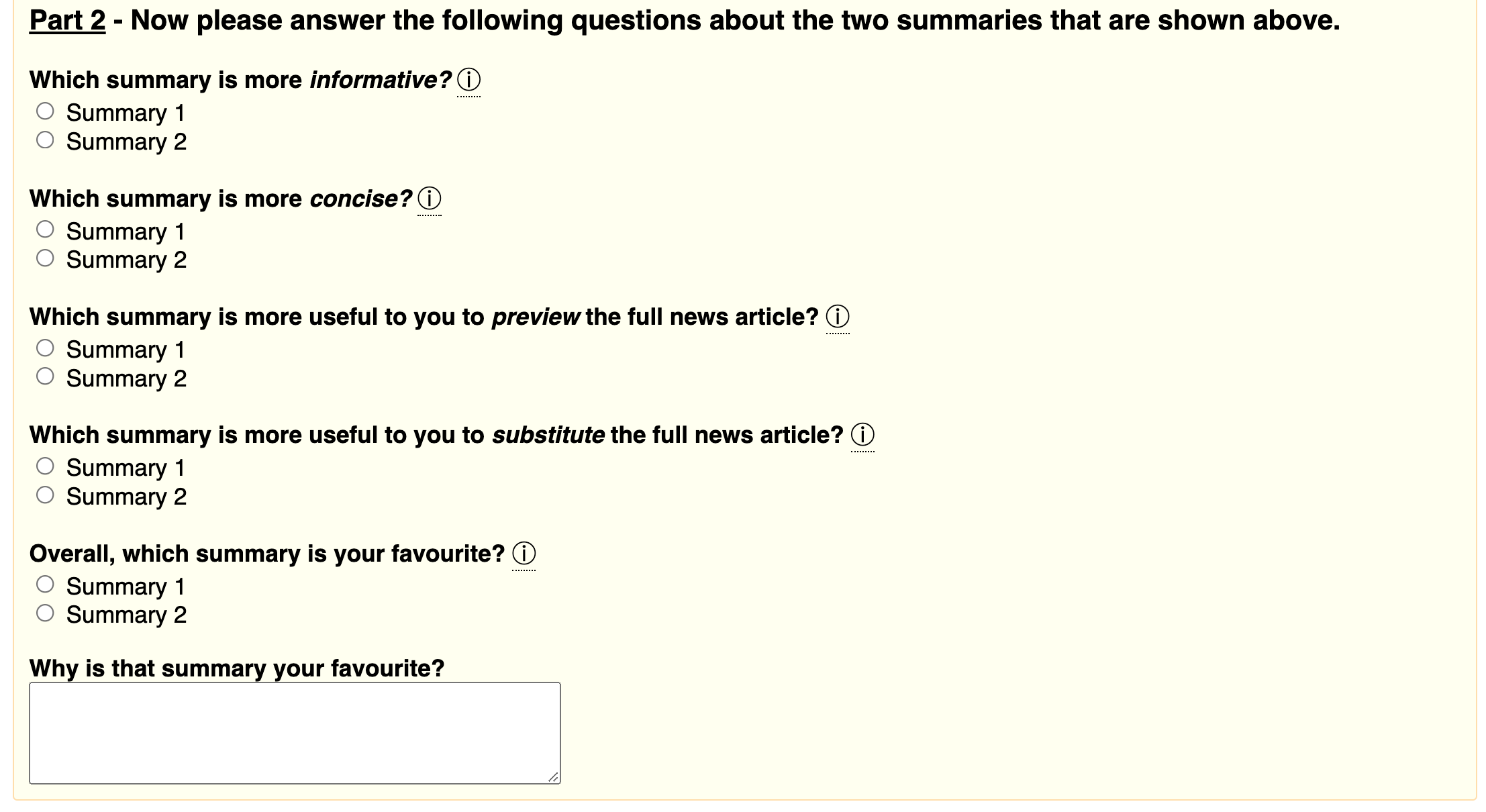}
    \end{subfigure}
    \caption{Example of human evaluation task. These are screen shots -- Summaries have a larger font in the actual task so they are more readable for workers.}
\label{fig:human_eval_q2_overview}
  \centering
\end{figure*}

\twocolumn
\subsection{Examples of generated summaries}

\begin{figure*}
    \centering
    \begin{subfigure}{\textwidth}
        \centering
        \includegraphics[width=\textwidth]{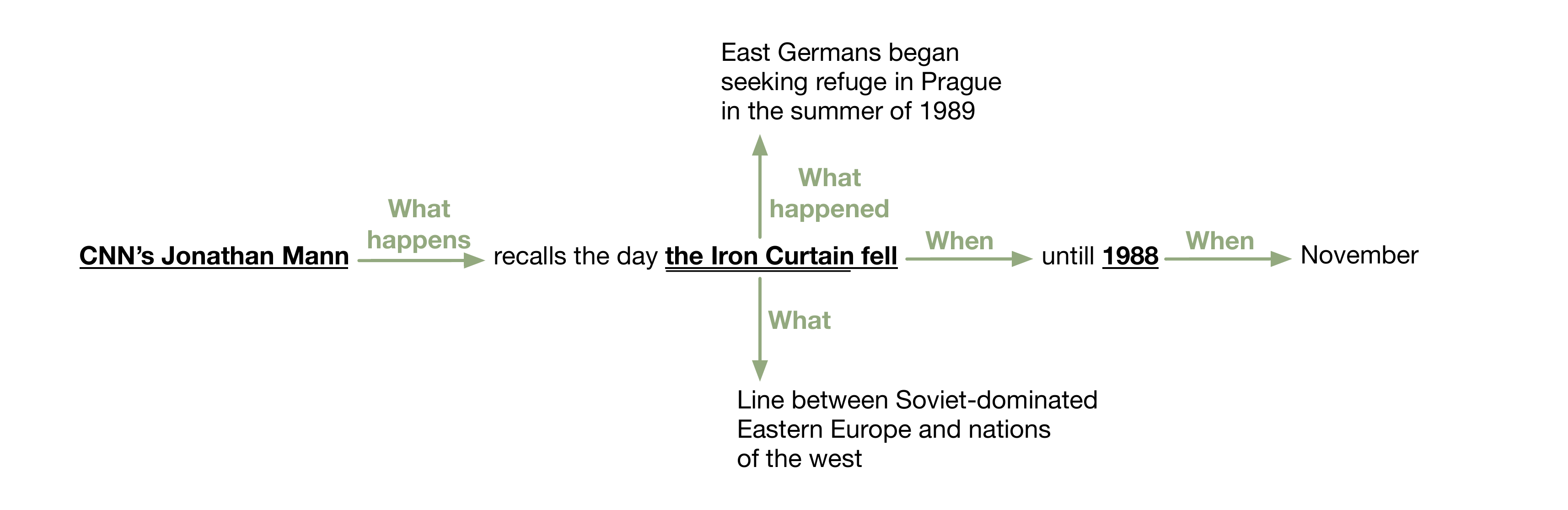}
        \caption{Human labeled summary.}
        \label{fig:human_eval_q2_example_1_hl}
    \end{subfigure}
    \begin{subfigure}{\textwidth}
        \centering
        \includegraphics[width=\textwidth]{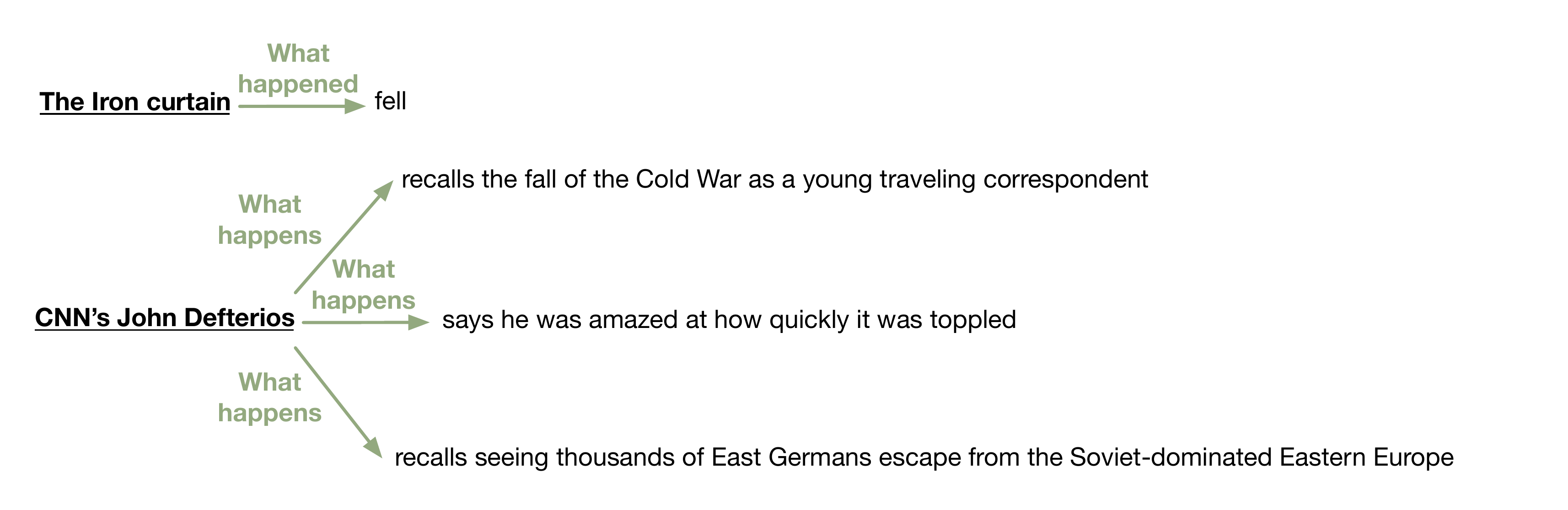}
        \caption{BART + Snorkel}
        \label{fig:human_eval_q2_example_1_snorkel}
    \end{subfigure}
    \begin{subfigure}{\textwidth}
        \centering
        \includegraphics[width=\textwidth]{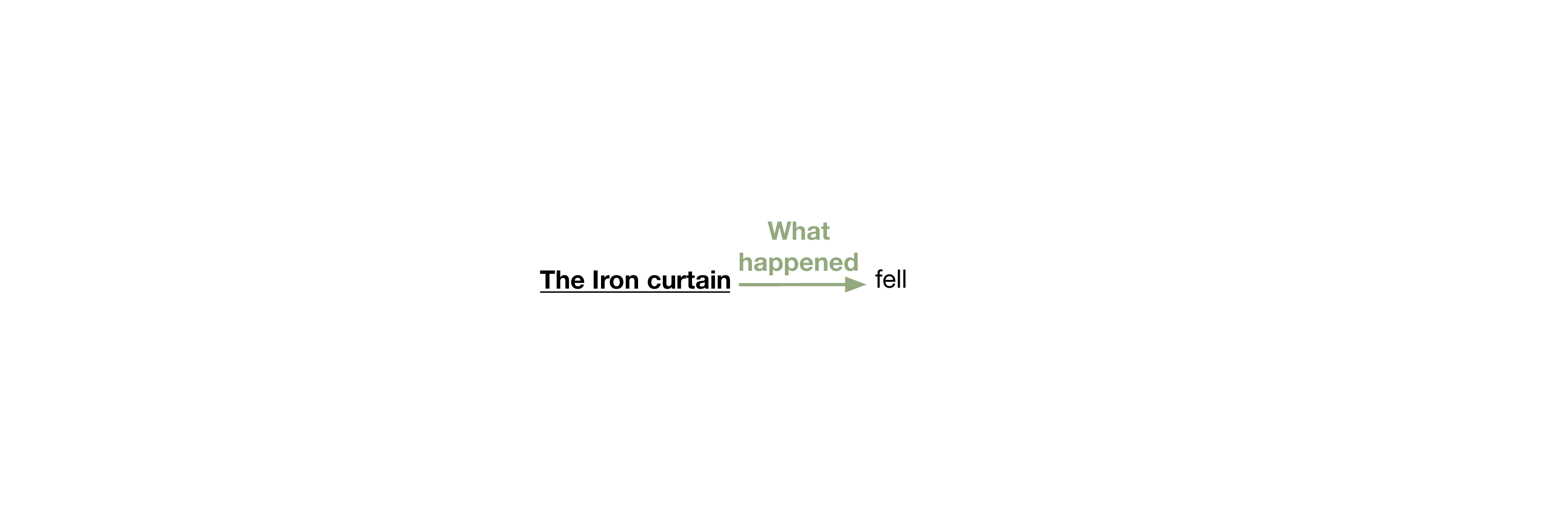}
        \caption{BART + \textsc{DieGIE++}}
        \label{fig:human_eval_q2_example_1_dygiepp}
    \end{subfigure}
    \caption{Examples of summaries with graphical elements generated by different methods.}
\label{fig:human_eval_q2_example_1}
  \centering
\end{figure*}

\begin{figure*}
    \centering
    \begin{subfigure}{\textwidth}
        \centering
        \includegraphics[width=\textwidth]{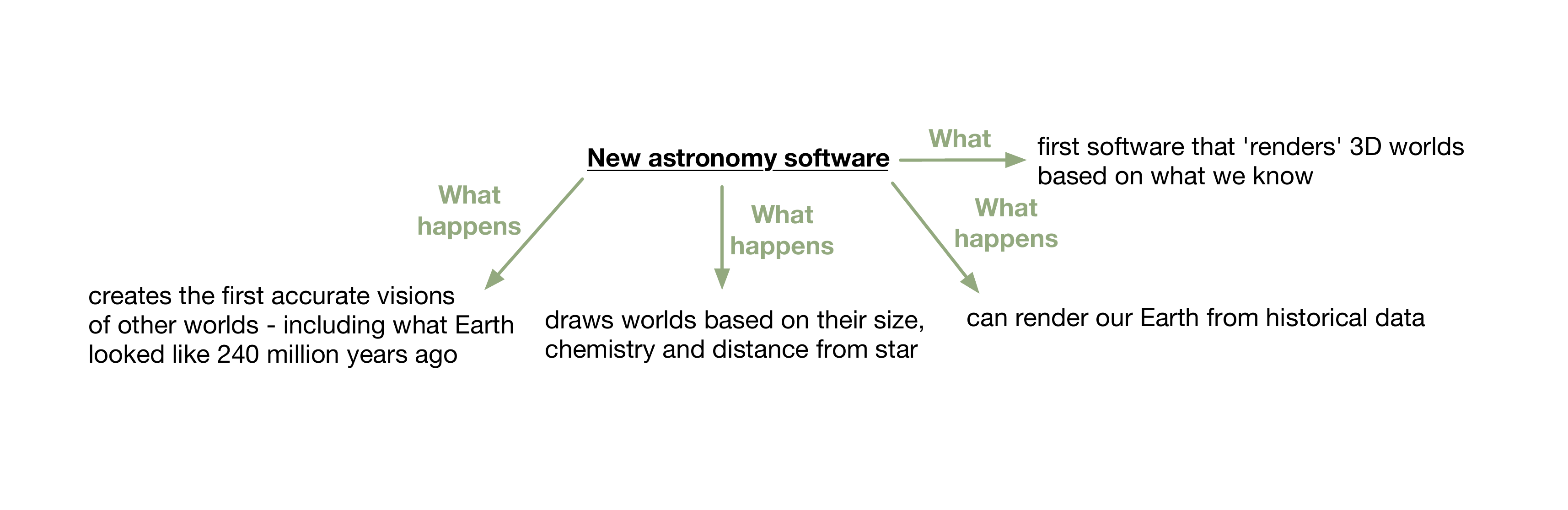}
        \caption{Human labeled summary.}
        \label{fig:human_eval_q2_example_2_hl}
    \end{subfigure}
    \begin{subfigure}{\textwidth}
        \centering
        \includegraphics[width=\textwidth]{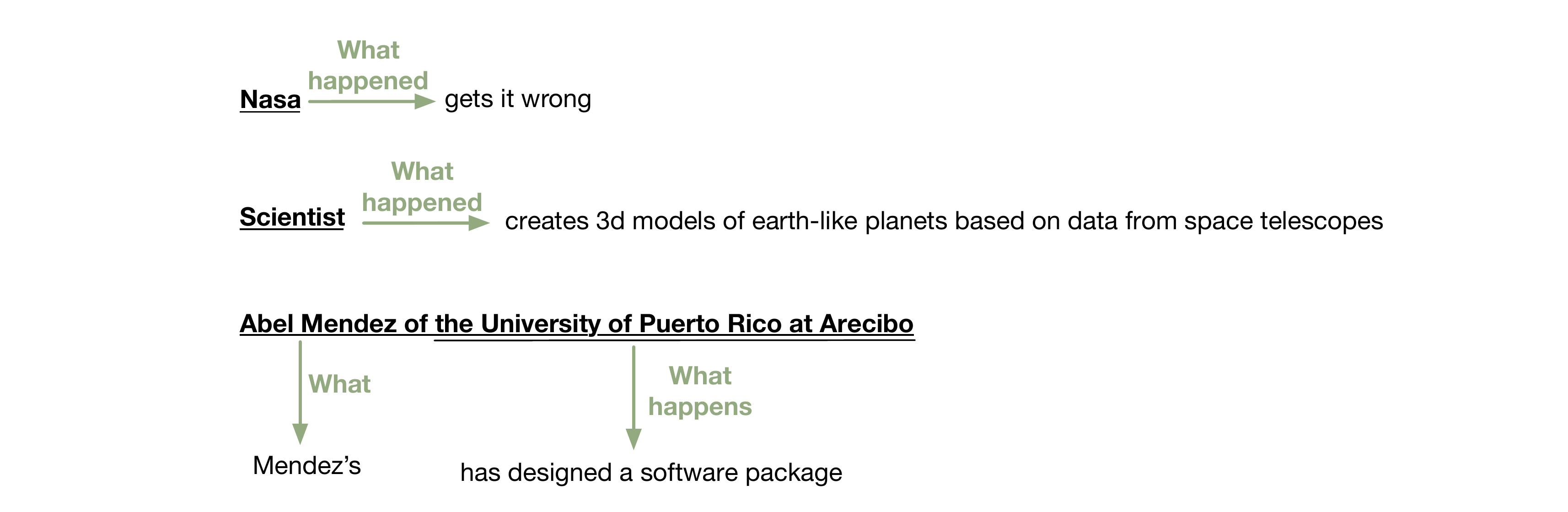}
        \caption{BART + Snorkel}
        \label{fig:human_eval_q2_example_2_snorkel}
    \end{subfigure}
    \begin{subfigure}{\textwidth}
        \centering
        \includegraphics[width=\textwidth]{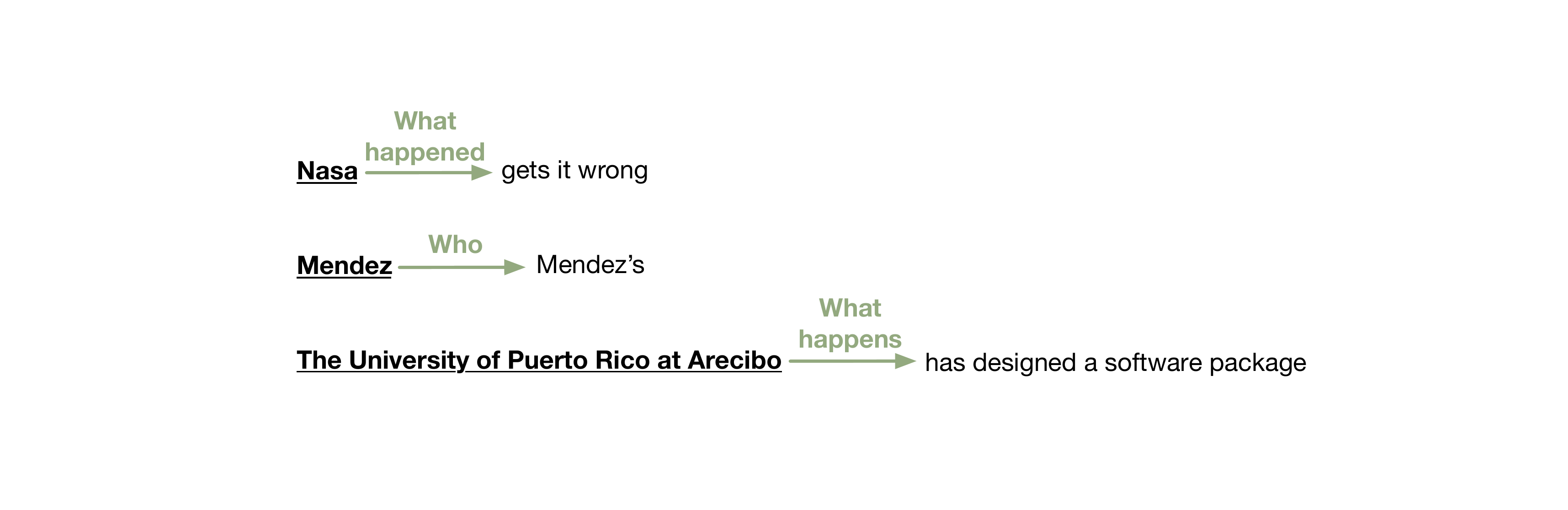}
        \caption{BART + \textsc{DieGIE++}}
        \label{fig:human_eval_q2_example_2_dygiepp}
    \end{subfigure}
    \caption{Examples of summaries with graphical elements generated by different methods.}
\label{fig:human_eval_q2_example_2}
  \centering
\end{figure*}

Figure~\ref{fig:human_eval_q2_example_1} and Figure~\ref{fig:human_eval_q2_example_2} give examples of outputs of summaries with graphical elements, generated by different methods.

\onecolumn

%!TEX root = ../main.tex

\section{Additional results automatic evaluation}
\label{sec:appendix_automatic_eval}
\onecolumn
\begin{table*}
\centering
\begin{tabular}{l rrrr}
\toprule
& \multicolumn{1}{c}{\textbf{Avg}} & \multicolumn{1}{c}{\textbf{A}} & \multicolumn{1}{c}{\textbf{B}} & \multicolumn{1}{c}{\textbf{Rel}} \\
\midrule
GT Abstract + Snorkel & $0.065 \pm 0.089$ & $0.238 \pm 0.175$ & $0.19 \pm 0.148$ & $0.582 \pm 0.256$ \\
GT Abstract + \textsc{DyGIE++} & $0.026 \pm 0.057$ & $0.161 \pm 0.204$ & $0.095 \pm 0.103$ & $0.409 \pm 0.262$  \\
\midrule
BART + Snorkel & $0.001 \pm 0.009$ & $0.086 \pm 0.132$ & $0.016 \pm 0.041$ & $0.535 \pm 0.221$  \\
T5 + Snorkel & $0.003 \pm 0.021$ & $0.091 \pm 0.134$ & $0.017 \pm 0.047$ & $0.543 \pm 0.224$  \\
\midrule
BART + \textsc{DyGIE++} & $0.000 \pm 0.005$ & $0.056 \pm 0.126$ & $0.017 \pm 0.047$ & $0.352 \pm 0.270$ \\
T5 + \textsc{DyGIE++} & $0.001 \pm 0.008$ & $0.065 \pm 0.132$ & $0.014 \pm 0.045$ & $0.345 \pm 0.259$  \\
\bottomrule
\end{tabular}
\caption{Jaccard scores entire pipeline.}
\label{tab:results_automatic_eval_pipeline_scores}
\end{table*}

\begin{table*}
\centering
\begin{tabular}{l rrr}
\toprule
 & \multicolumn{3}{c}{\textbf{Soft Binary}}   \\
\cmidrule(lr){2-4}
& \multicolumn{1}{c}{\textbf{P}} & \multicolumn{1}{c}{\textbf{R}} & \multicolumn{1}{c}{\textbf{F1}}  \\
\midrule
GT Abstract + Snorkel & $0.342 \pm 0.241$ & $0.262 \pm 0.200$ & $  0.286 \pm 0.202$ \\
GT Abstract + \textsc{DyGIE++} & $0.237 \pm 0.303$ & $0.102 \pm 0.139$ & $0.132 \pm 0.162$ \\
\midrule
BART + Snorkel & $0.239 \pm 0.219$ & $0.173 \pm 0.164$ & $0.191 \pm 0.170$   \\
T5 + Snorkel & $0.250 \pm 0.247$ & $0.179 \pm 0.195$ & $0.198 \pm 0.194$  \\
\midrule
BART + \textsc{DyGIE++} & $0.145 \pm 0.268$ & $0.056 \pm 0.101$ & $0.075 \pm 0.129$  \\
T5 + \textsc{DyGIE++} & $0.137 \pm 0.261$ & $0.055 \pm 0.102$ & $0.072\pm 0.128$  \\
\bottomrule
\end{tabular}
\caption{We compute an additional soft matching score, where we count whether or not summary triples have overlap with the ground truth triples.}
\label{tab:results_automatic_eval_pipeline_soft}
\end{table*}

\begin{figure*}
    \centering
    \begin{subfigure}{0.45\textwidth}
        \centering
        \includegraphics[width=\textwidth]{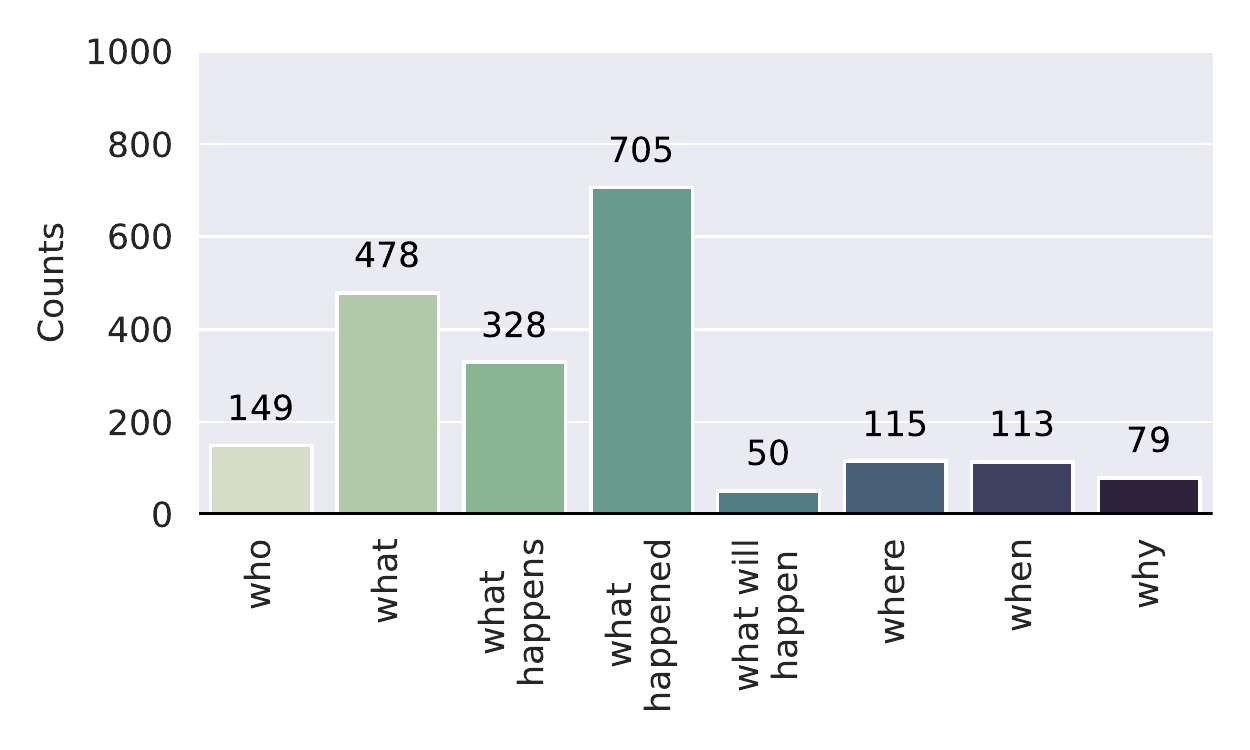}
        \caption{Human labeled test set. Counts are averaged over the average number of annotations per document.}
    \end{subfigure}
    
    \begin{subfigure}{0.45\textwidth}
        \centering
        \includegraphics[width=\textwidth]{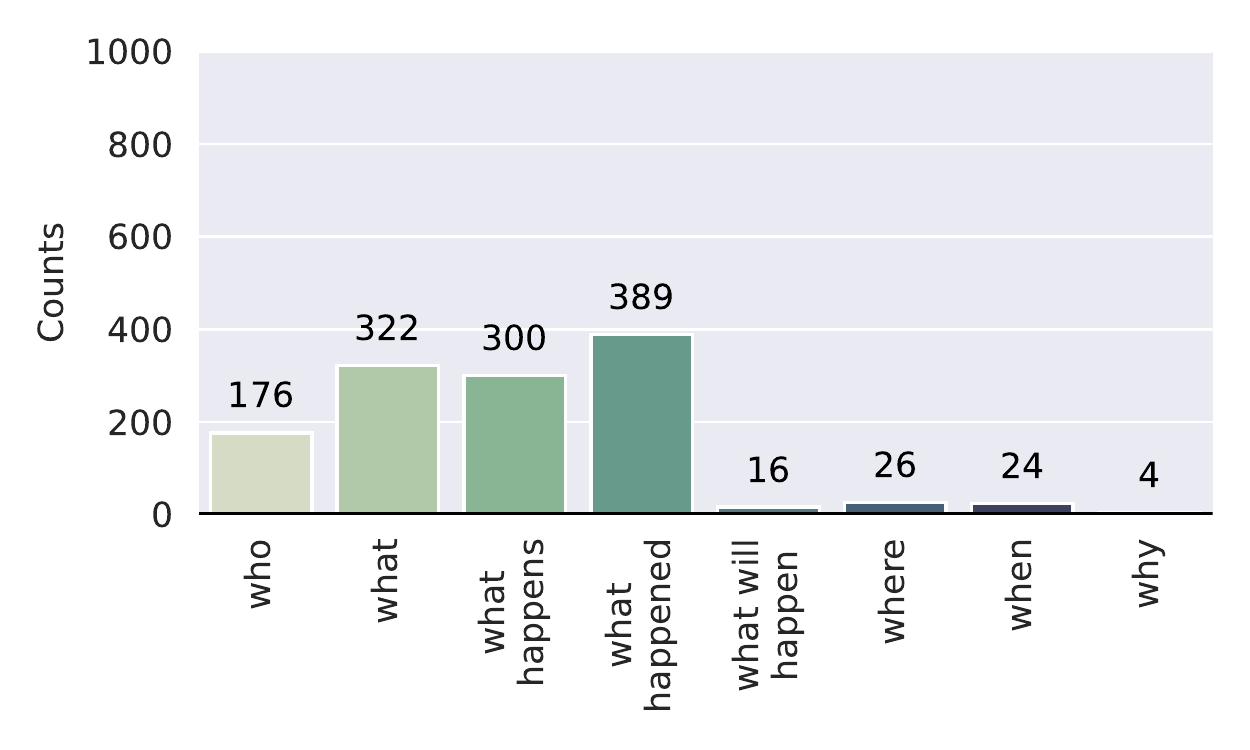}
        \caption{GT Abstract + Snorkel}
    \end{subfigure}%
        \begin{subfigure}{0.45\textwidth}
        \centering
        \includegraphics[width=\textwidth]{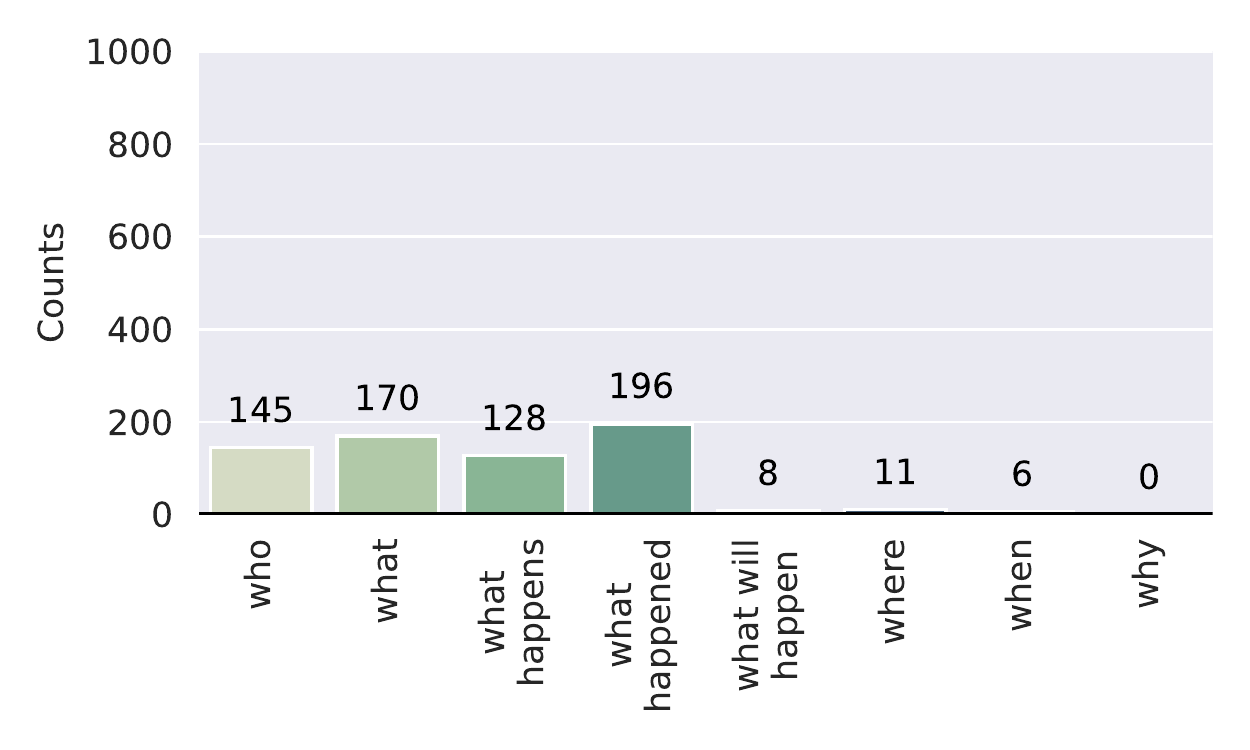}
        \caption{GT Abstract + \textsc{DyGIE++}}
    \end{subfigure}
    
    \begin{subfigure}{0.45\textwidth}
        \centering
        \includegraphics[width=\textwidth]{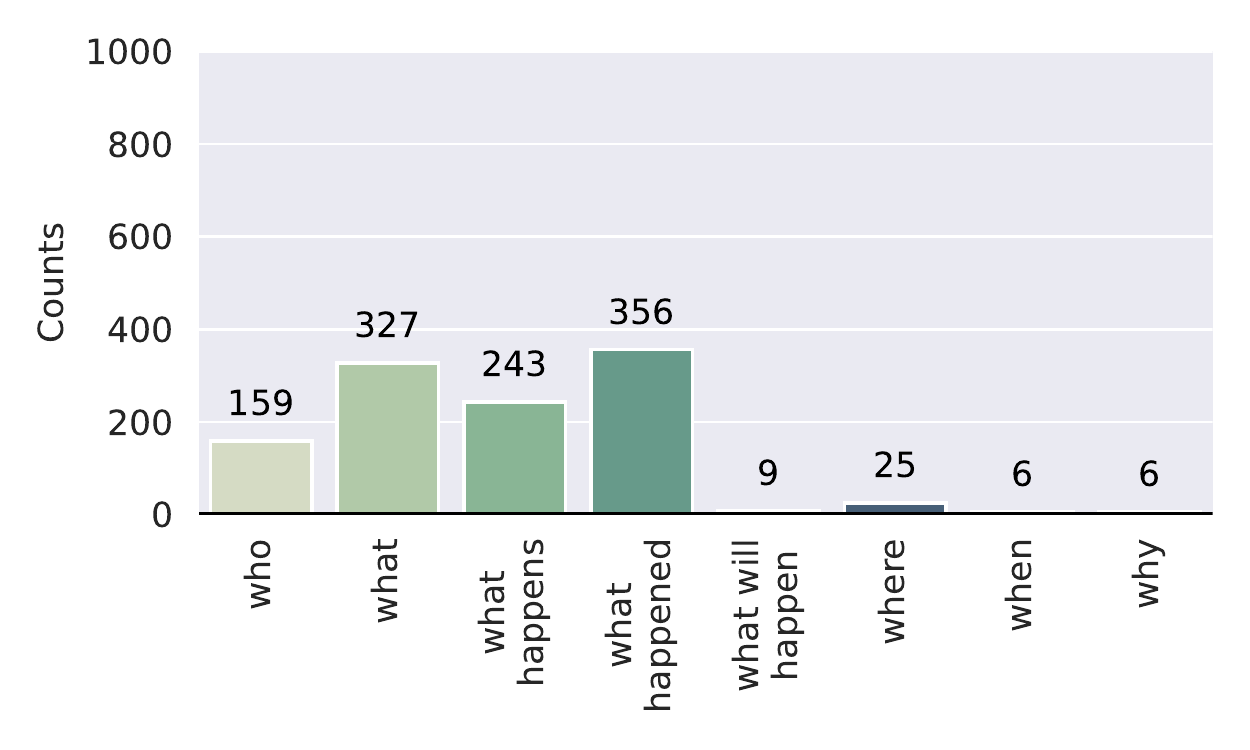}
        \caption{BART + Snorkel}
    \end{subfigure}%
    \begin{subfigure}{0.45\textwidth}
        \centering
        \includegraphics[width=\textwidth]{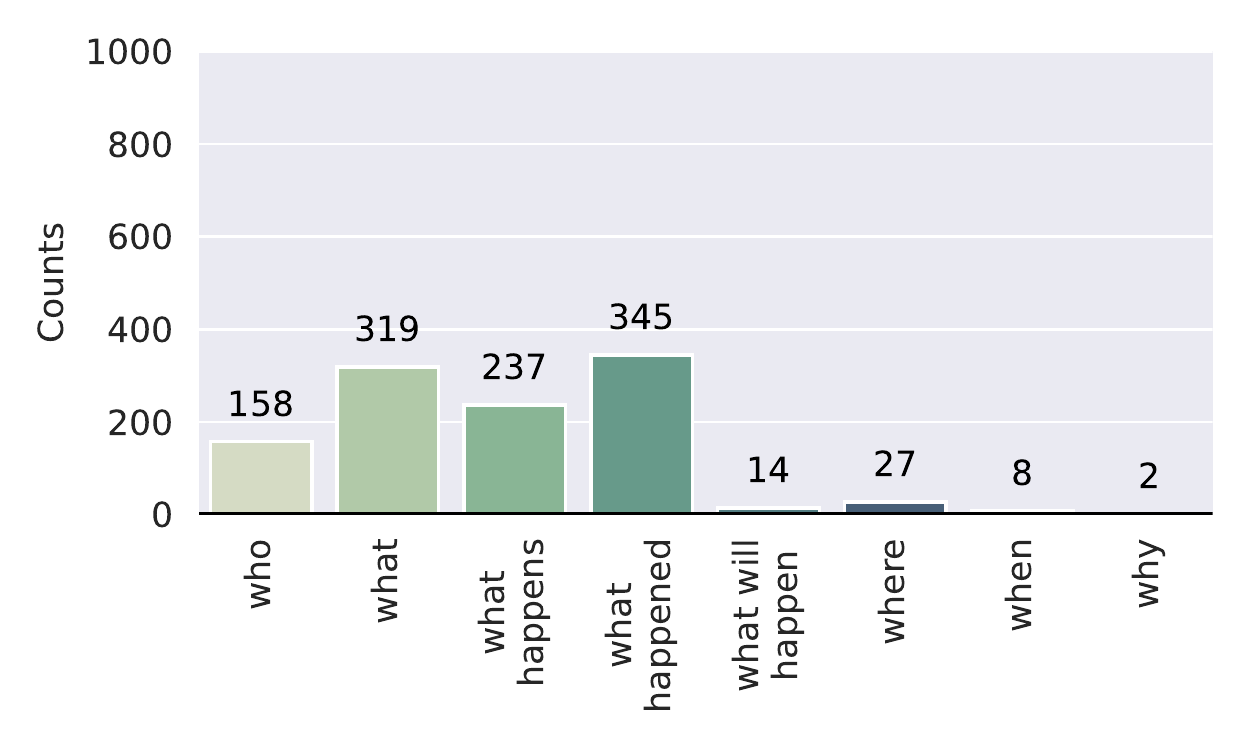}
        \caption{T5 + Snorkel}
    \end{subfigure}
    
    \begin{subfigure}{0.45\textwidth}
        \centering
        \includegraphics[width=\textwidth]{figures/bart_snorkel_relation_counts.pdf}
        \caption{BART + \textsc{DyGIE++}}
    \end{subfigure}%
    \begin{subfigure}{0.45\textwidth}
        \centering
        \includegraphics[width=\textwidth]{figures/t5_snorkel_relation_counts.pdf}
        \caption{T5 + \textsc{DyGIE++}}
    \end{subfigure}
    \caption{Histograms of relation counts for several baselines.}
\label{fig:hists_relations_counts_baselines}
  \centering
\end{figure*}

\begin{figure*}
    \centering
    \begin{subfigure}{0.45\textwidth}
        \centering
        \includegraphics[width=\textwidth]{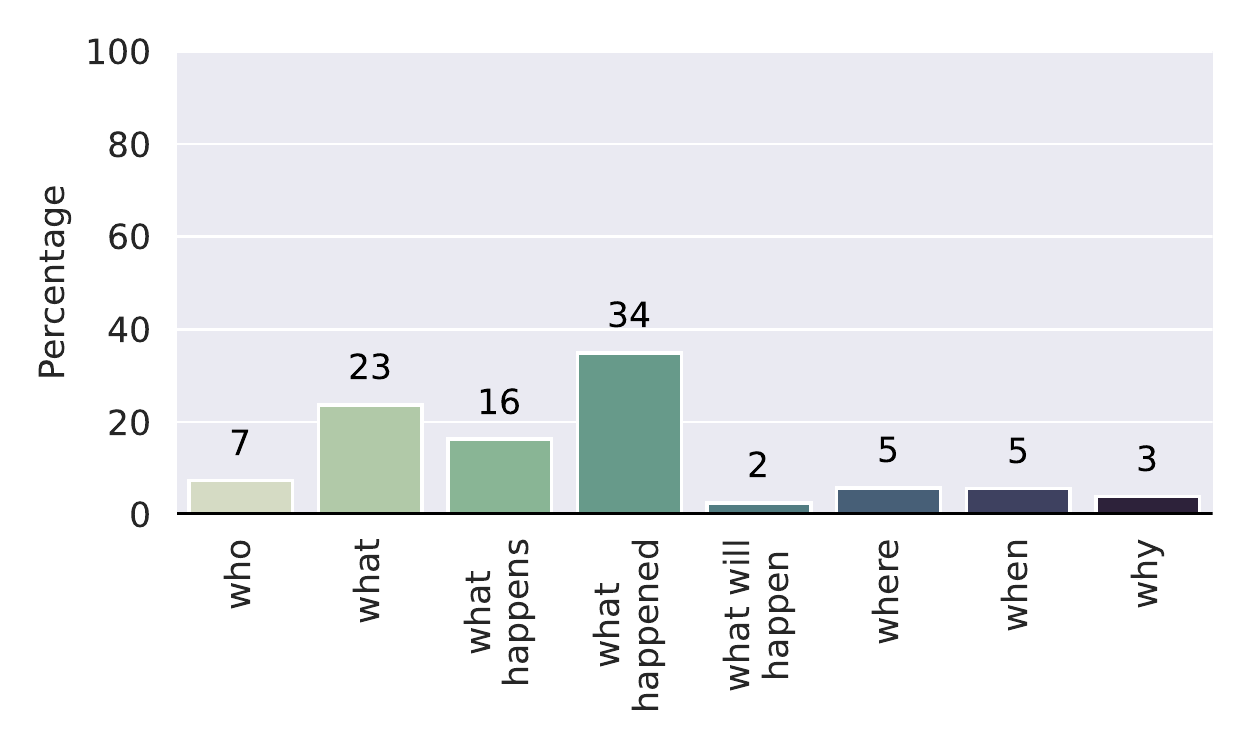}
        \caption{Human labeled test set}
    \end{subfigure}
    
    \begin{subfigure}{0.45\textwidth}
        \centering
        \includegraphics[width=\textwidth]{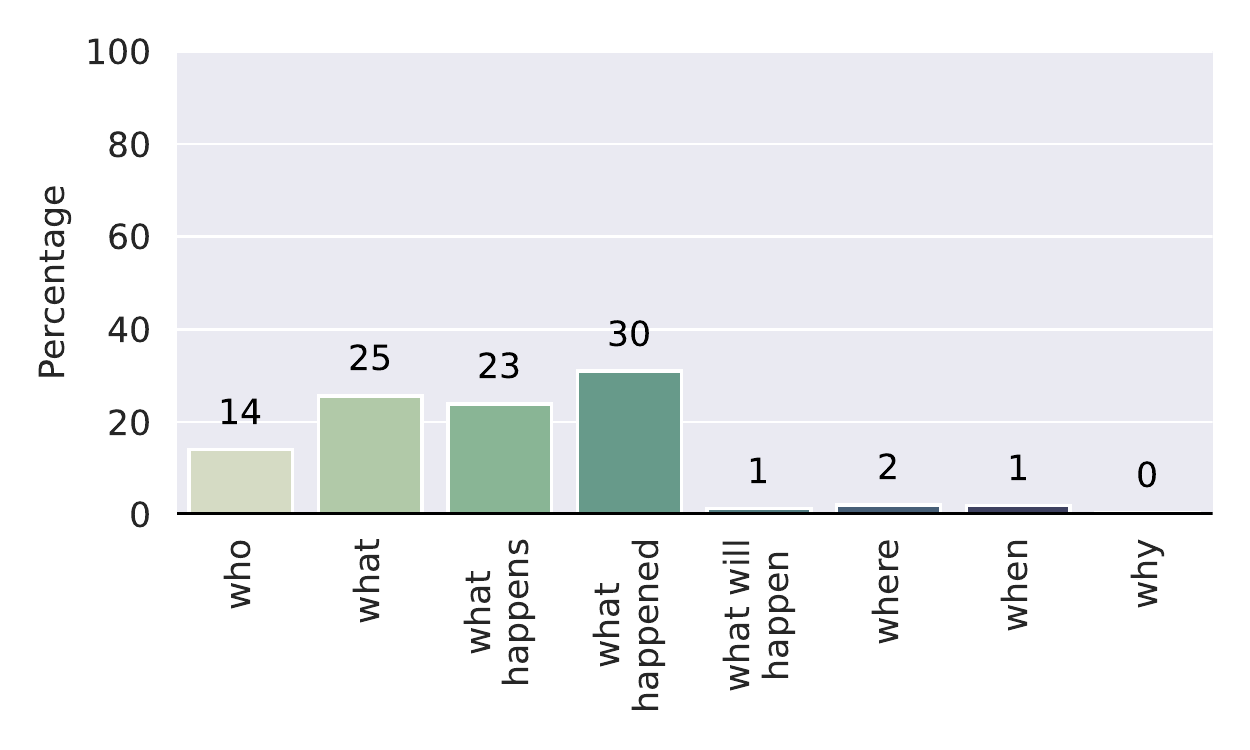}
        \caption{GT Abstract + Snorkel}
    \end{subfigure}%
        \begin{subfigure}{0.45\textwidth}
        \centering
        \includegraphics[width=\textwidth]{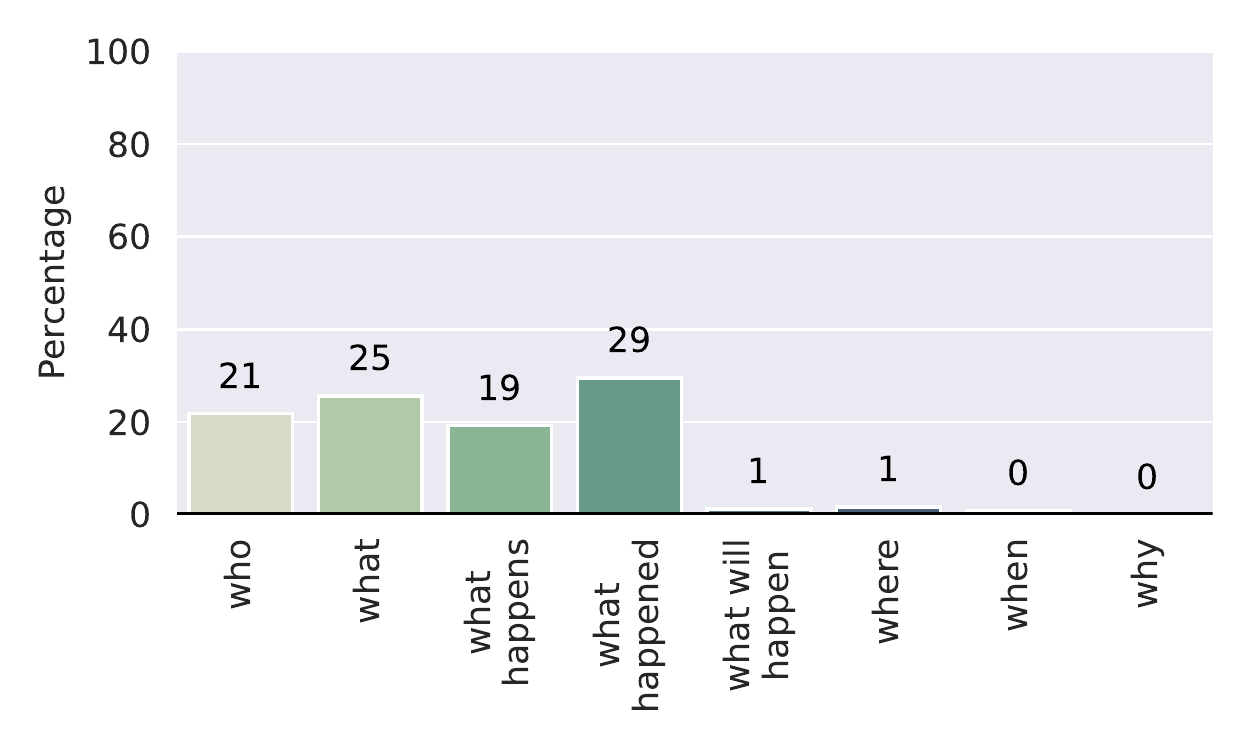}
        \caption{GT Abstract + \textsc{DyGIE++}}
    \end{subfigure}
    
    \begin{subfigure}{0.45\textwidth}
        \centering
        \includegraphics[width=\textwidth]{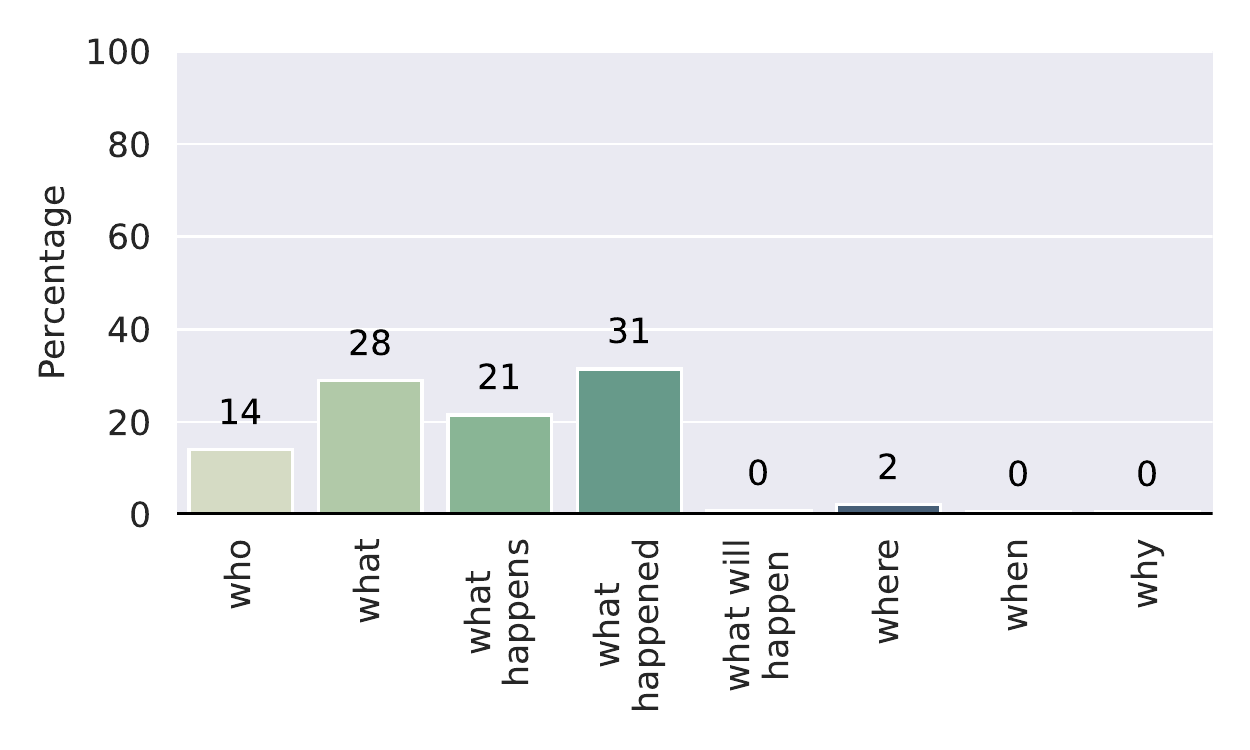}
        \caption{BART + Snorkel}
    \end{subfigure}%
    \begin{subfigure}{0.45\textwidth}
        \centering
        \includegraphics[width=\textwidth]{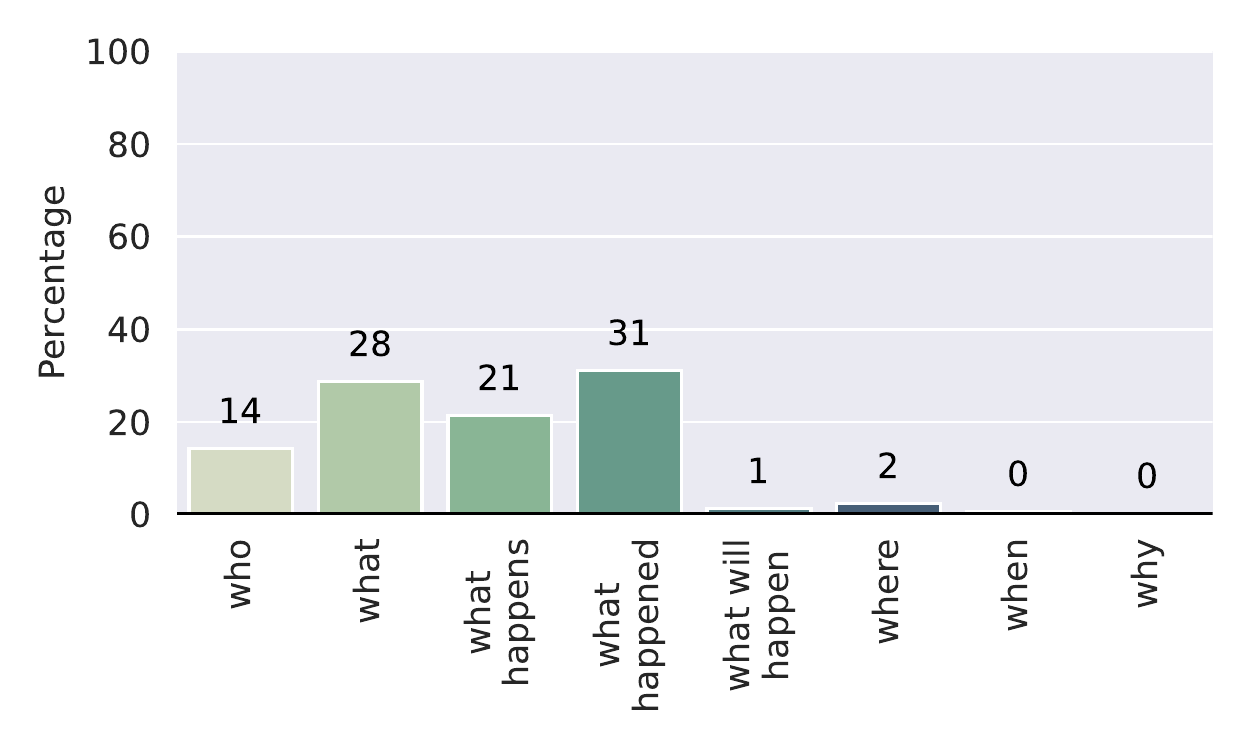}
        \caption{T5 + Snorkel}
    \end{subfigure}
    
    \begin{subfigure}{0.45\textwidth}
        \centering
        \includegraphics[width=\textwidth]{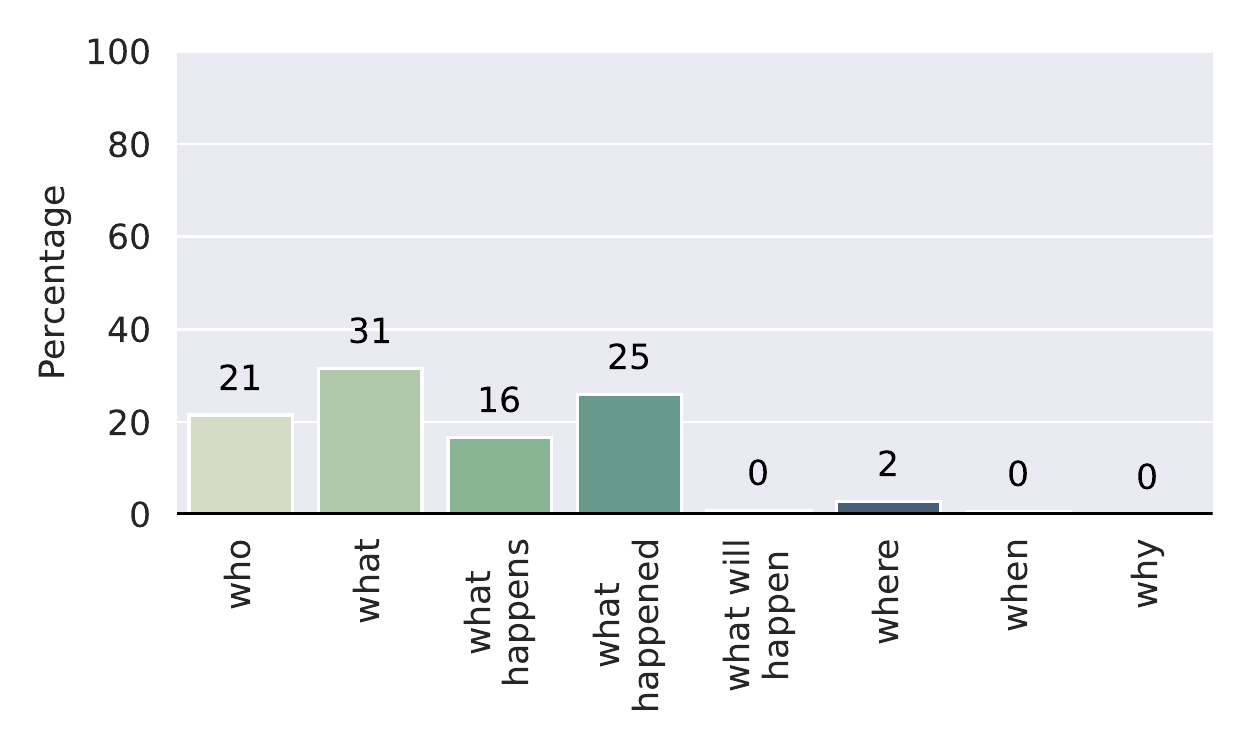}
        \caption{BART + \textsc{DyGIE++}}
    \end{subfigure}%
    \begin{subfigure}{0.45\textwidth}
        \centering
        \includegraphics[width=\textwidth]{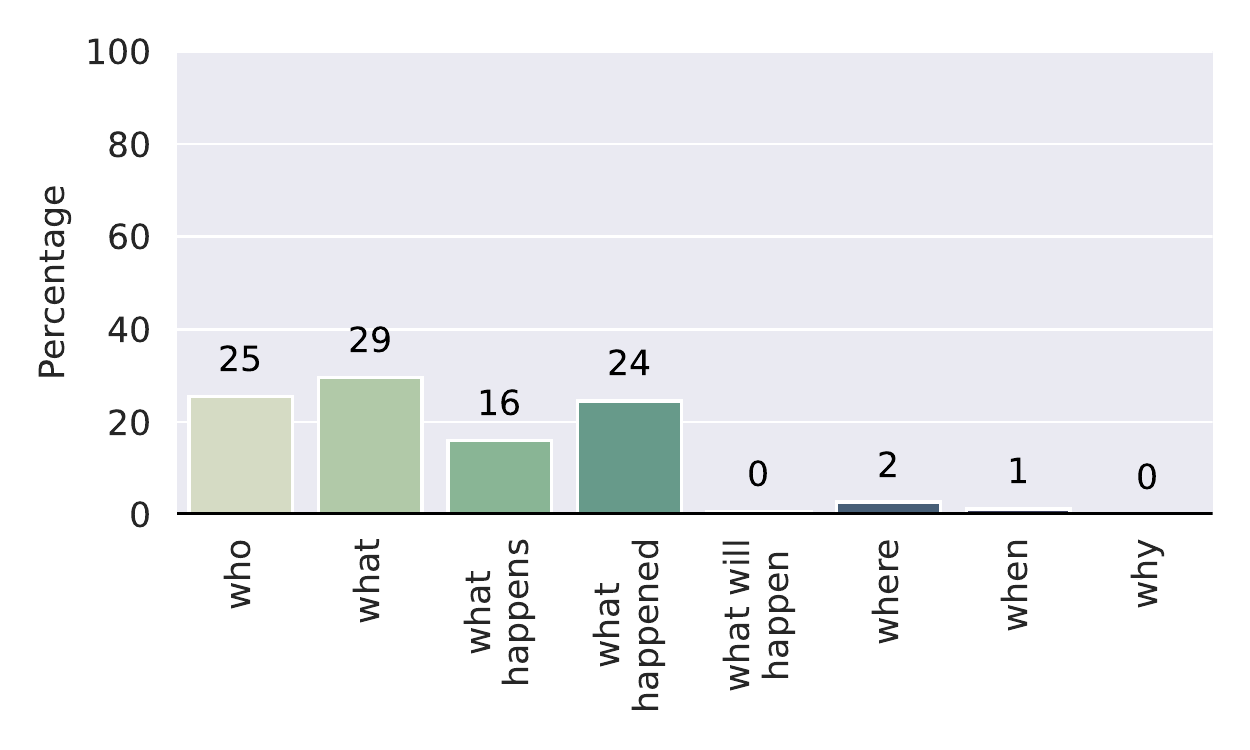}
        \caption{T5 + \textsc{DyGIE++}}
    \end{subfigure}
    
    \caption{Histograms of relation percentages for several baselines.}
\label{fig:hists_relations_percentage_baselines}
  \centering
\end{figure*}

\end{document}